\documentclass[reprint, amsmath, amssymb, aps, prl, superscriptaddress]{revtex4-2}
\usepackage[version=3]{mhchem} 

\usepackage{graphicx}
\usepackage{dcolumn}
\usepackage{bm}
\usepackage{float}
\usepackage{hyperref}
\usepackage[dvipsnames, table]{xcolor}
\usepackage{gensymb}
\usepackage{pbox}
\usepackage{changes}
\usepackage{mathtools}
\usepackage{algpseudocode}

\begin{document}

\author{F\'elix Therrien}
\affiliation
{University of Toronto, Department of Physical \& Environmental Science}
\affiliation
{University of Toronto, Department of Electrical and Computer Engineering}


\author{Edward H. Sargent}
\affiliation
{University of Toronto, Department of Electrical and Computer Engineering}

\author{Oleksandr Voznyy}
\affiliation
{University of Toronto, Department of Physical \& Environmental Science}
\email{o.voznyy@utoronto.ca}

\title{Using GNN property predictors as molecule generators}


\begin{abstract}


Graph neural networks (GNNs) have emerged as powerful tools to accurately predict materials and molecular properties in computational discovery pipelines. 
 In this article, we exploit the invertible nature of these neural networks to directly generate molecular structures with desired electronic properties. Starting from a random graph or an existing molecule, we perform a gradient ascent while holding the GNN weights fixed in order to optimize its input, the molecular graph, towards the target property. Valence rules are enforced strictly through a judicious graph construction. The method relies entirely on the property predictor; no additional training is required on molecular structures. 
We demonstrate the application of this method by generating molecules with specific DFT-verified energy gaps and octanol-water partition coefficients (logP). Our approach hits target properties with rates comparable to or better than state-of-the-art generative models while consistently generating more diverse molecules.


\end{abstract}

\maketitle

One of the ultimate goals of computational materials science is to be able to rapidly uncover material structures and compositions with specific properties \cite{zunger2018inverse}. This is particularly true in the fields of pharmacy and  materials for energy and sustainability where time is of the essence. In the past decades, computational materials discovery has been achieved by going through large databases of existing materials and computing properties from first principles using methods such as density functional theory (DFT) and molecular dynamics \cite{lu2021computational, therrien2021metastable}. These methods have proven to be successful in many cases \cite{gorai2017computationally, garrity2018high, curtarolo2013high, alberi20182019}, but suffer from two main limitations: (1) they are computationally expensive and (2) they cover a small subspace of all possible materials.

In an effort to alleviate the first problem, machine learning (ML) based property prediction methods have become an integral part of materials science \cite{schmidt2019recent, butler2018machine}. Various models exploit different materials representations (e.g., graph, fingerprints) \cite{reiser2022graph, wigh2022review}, model architectures (e.g., neural networks, random forest) \cite{wei2019machine} and datasets (experimental or computed) and their accuracy has been steadily increasing, often competing with that of DFT \cite{schmidt2019recent}. The success and adoption of these models is due largely to the powerful tools developed by the ML and data science communities (such as Pytorch\cite{paszke2019pytorch}, Tensorflow \cite{tensorflow2015-whitepaper}, Pandas \cite{reback2020pandas}, etc.). However, despite their promising performance on benchmark datasets, ML property predictors still suffer from poor generalizability \cite{li2023critical}, exhibiting much lower performance on out-of-distribution data, i.e., materials that are different from what they have been trained on.

Materials and molecule generation can alleviate the second limitation: it can theoretically explore the full space of all possible materials. Traditionally this has been done using minima hopping \cite{goedecker2004minima}, metadynamics\cite{martovnak2003predicting} and evolutionary approaches \cite{glass2006uspex, jensen2019graph, oganov2019structure}, but machines can also learn to \textit{generate} realistic materials \cite{lu2021computational}. The goal is no longer to predict properties but to predict realistic structures \cite{sanchez2018inverse}. Materials and molecular generation using ML is a rapidly evolving field with a large body of recent methods including variational autoencoders \cite{jin2018junction, gomez2018automatic, eckmann2022limo}, flow-based models \cite{bengio2021flow, roy2023goal}, diffusion models \cite{vignac2022digress, xu2022geodiff}, models based on reinforcement learning (RL) \cite{guimaraes2017objective, you2018graph} and many others \cite{kong2023molecule, nigam2022parallel}.

\begin{figure*}
\includegraphics[width=1.0\linewidth]{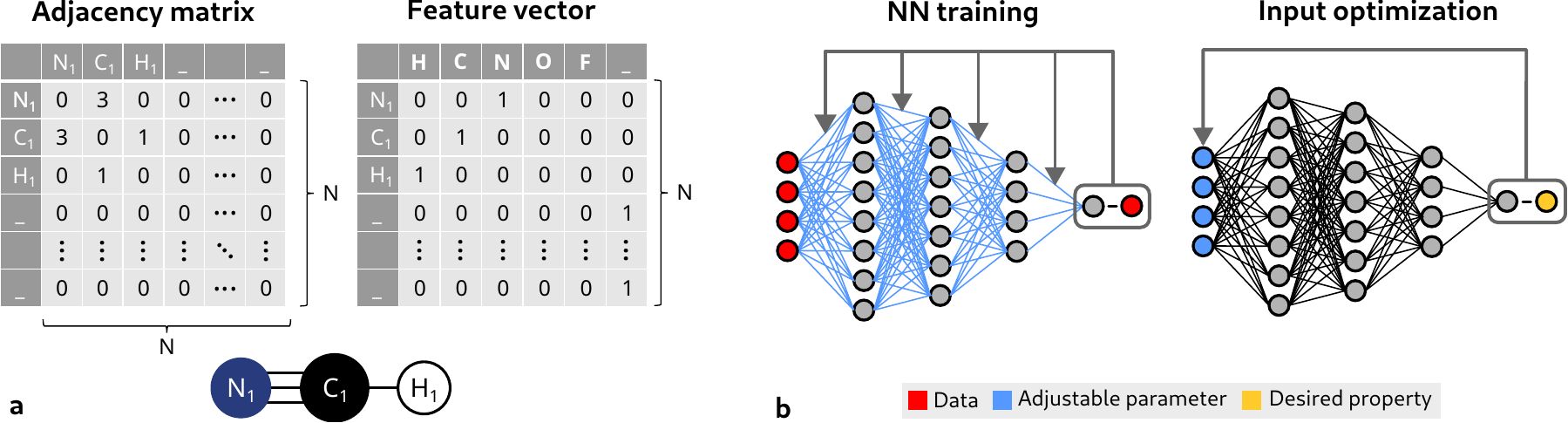}
\caption{\label{fig:intro}
a) Molecular representation for this work using an HCN molecule as an example. b) A visual representation of a typical training process for a neural network in comparison to an input optimization scheme.}
\end{figure*}

In this paper, we exploit one of the most important and fundamental features of neural networks, their differentiability, to directly optimize a target property with respect to the graph representation itself starting from a pre-trained predictive model. This concept sometimes termed \textit{gradient ascent}, or \textit{input optimization}, has been used extensively in other fields \cite{trabucco2022design, linder2021fast} and a similar idea has been applied to molecular generation with SELFIES \cite{shen2021deep}. Here we apply input optimization to molecular GNNs. We describe how carefully constraining the molecular representation makes this ``naive'' approach possible and show that it can generate molecules with requested properties as verified with density functional theory and empirical models. It does so with comparable or better performance than existing methods while consistently generating the most diverse set of molecules.

\section{Results}

\subsection{Rationale and workflow overview}

Our method can technically be applied to any GNN architecture that uses molecular graphs. To train our GNN we use (1) an explicit representation of the adjacency matrix where non-zero elements are the bond orders and (2) a feature matrix that contains a one-hot representation of the atoms. These two matrices fully describe the graph and contain exactly the same information as a SMILES string.  Since all functions in the GNN have well-defined gradients (allowing it to be trained in the first place), the adjacency matrix and the feature vector can be optimized through a gradient descent with respect to a target property as illustrated in Figure~\ref{fig:intro}. This is termed \textit{gradient ascent}--although it does not change the direction of the gradients but rather, the variable with respect to which they are taken. This approach could be seen as a "naive" way to tackle the problem of conditional molecular and materials generation; unconstrained, it would lead to meaningless results that do not follow the basic structures of the adjacency matrix (e.g., its symmetry) and the feature matrix (e.g., one non-zero element per line). The major contribution of this paper is to enforce structural and chemical rules such that optimized inputs can only be valid molecules allowing for direct optimization into graph space.

The adjacency matrix is constructed from a weight vector $\mathbf{w_{adj}}$ containing $\frac{N^2-N}{2}$ elements. These elements are squared and populated in an upper triangular matrix with zeros on the main diagonal. The resulting matrix is then added to its transpose to obtain a positive symmetric matrix with zero trace. Elements of the matrix are then rounded to the nearest integer to obtain the adjacency matrix. The key element here is that the adjacency matrix needs to have non-zero gradients with respect to $\mathbf{w_{adj}}$, which is not the case when using a conventional rounding half-up function. To alleviate this problem we used a sloped rounding function, 
\begin{equation}
[x]_{\text{sloped}} = [x] + a(x - [x]) 
\end{equation}
where $[x]$ is the conventional rounding half-up function and $a$ is an adjustable hyper-parameter. These steps guarantee that only valid near-integer filled adjacency matrices are constructed. However, it does not take into account any chemistry: atoms are allowed to form as many bonds as there are rows in the adjacency matrix. To avoid that, we use two strategies: (1) we penalize valence (sum of bond orders) of more than 4 through the loss function and (2) we do not allow gradients in the direction of higher number of bonds when the valence is already 4.

The feature vector, on the other hand, is constructed directly from the adjacency matrix. The idea is to \textit{define} the atoms from their valence, i.e., the sum of their bond orders. For example, a node with four edges of value one, or, in other words, an atom forming four single bonds, would be defined as a carbon atom, a node forming one double bond would be defined as an oxygen atom, a node forming a double bond and a single bond would be defined as a nitrogen atom, etc.  In terms of matrices, this means that the sum of a row (or columns) of the adjacency matrix defines the element associated with that row (or column).

More details on how constraints on the adjacency matrix and feature vectors are implemented can be found in the supplemental information.

\subsection{Energy gap targeting}
\begin{figure}
\includegraphics[width=1.0\linewidth]{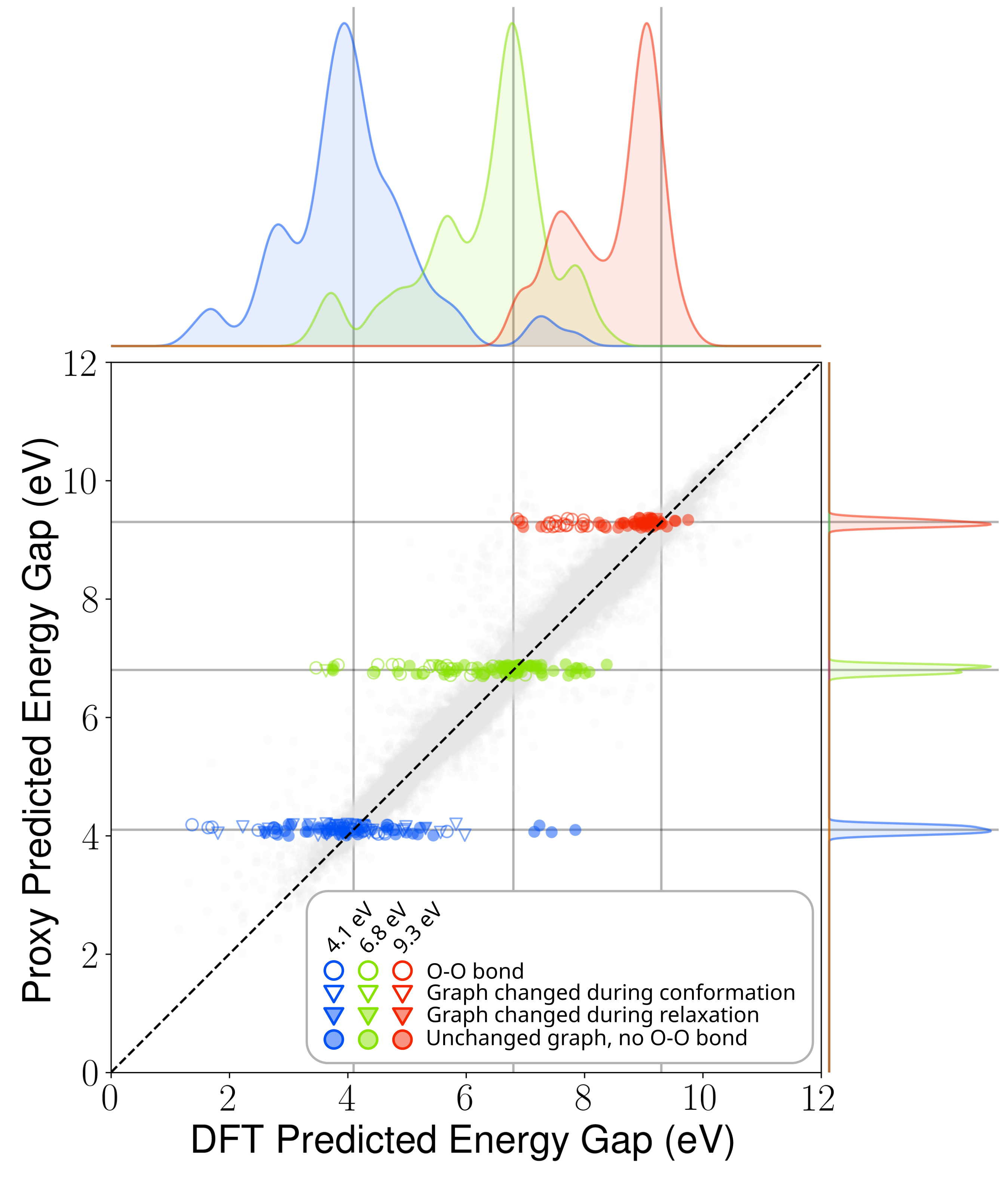}
\caption{\label{fig:gen}
Generated molecules HOMO-LUMO gap DFT and proxy predictions. Generated molecules are overlaid on the proxy model performance on the QM9 dataset (test + train).}
\end{figure}
\begin{table*}
\begin{ruledtabular}
\caption{\label{tab:gap} Comparison of generated molecules from our procedure (DIDgen) with those generated with a state-of-the-art genetic algorithm (JANUS) \cite{nigam2022parallel}. The comparison metrics are the number of DFT calculations ($n_\text{calcs}$), the number of molecules that are within 0.5~eV of the target ($n_{\pm0.5}$), the mean absolute error (MAE) to the target property and the diversity (Div.) of the generated molecules that are within 0.5~eV of the property.}
\begin{tabular}{c|cccc|cccc|cccc}
Method & \multicolumn{4}{c|}{4.1~eV} & \multicolumn{4}{c|}{6.8~eV} & \multicolumn{4}{c}{9.3~eV} \\
 & $n_\text{calcs}$ & $n_{\pm0.5}$ & MAE & Div. & $n_\text{calcs}$ & $n_{\pm0.5}$ & MAE & Div. & $n_\text{calcs}$ & $n_{\pm0.5}$ & MAE & Div.  \\
\colrule
QM9                     &       & 3.4\%         &               & 0.87          &       & 24\%          &               & 0.89          &       & 6.0\% &       & 0.84  \\
\colrule
JANUS (DFT)             & 197   & 24 (12.2\%)            & 0.96          & 0.79          & 392   & 42 (10.7\%)            & 0.92          & 0.80          & 484   & 26 (5.4\%)    & 1.28  & 0.81  \\
JANUS (Proxy)           & 100   & 36            & 1.05          & 0.86          & 100   & 46            & \textbf{0.80} & 0.82          & 100   & \textbf{37}    & 1.24  & 0.81  \\
DIDgen                  & 100   & \textbf{46}            & \textbf{0.81} & \textbf{0.91} & 100   & \textbf{50}   & 0.83          & \textbf{0.90} & 100   & 34    & \textbf{0.83}  & \textbf{0.83}  \\
\end{tabular}
\end{ruledtabular}
\end{table*}

The first task is to generate molecules with a specific energy gap $\mu$ between their highest occupied molecular orbital (HOMO) and their lowest unoccupied molecular orbital (LUMO). The HOMO-LUMO gap is a particularly interesting property because it is relevant to a number of applications, it is relatively expensive to compute and it is available in many databases \cite{ramakrishnan2014quantum, wu2018moleculenet, nakata2017pubchemqc, hachmann2011harvard}. Furthermore, the ability to generate molecules with a specific emission wavelength is of special interest to our own research on the discovery of efficient blue organic light emitting diodes (OLED) materials \cite{}.

%
We trained a simple GNN on the QM9 dataset \cite{ramakrishnan2014quantum} and used our direct inverse design procedure to generate 100 molecules with HOMO-LUMO gaps within 10~meV of 3 different target values: the first percentile of QM9 energy gaps, 4.1~eV, the median, 6.8~eV, and the 99$^\text{th}$ percentile, 9.3~eV. As an additional soft target, we aimed for compositions close to the average composition in QM9. This helps guide the generation towards molecules that the proxy can predict better. 

The results are illustrated in Figure~\ref{fig:gen}. Note that all generated molecules are within the requested range according to the proxy model \textit{by construction}: generation stops when that criterion is met. The DFT-calculated energy gap, on the other hand, is distributed around the requested property with relatively small overlap between different targets. 

The generated molecules predictions are overlaid on the predictions for the QM9 dataset. It is apparent that the proxy model's performance is significantly worse on generated molecules than on the test set. If the model was generalizing perfectly, we should expect the performance to be similar to that of the test set, MAE=0.12eV, rather than the observed performance of about 0.8~eV. This highlights the importance of benchmarking generating schemes on DFT-confirmed properties, not solely on ML predicted properties.

%
We compared our method to JANUS \cite{nigam2022parallel} an ML enhanced state-of-the-art genetic algorithm that was recently tested against several materials generations schemes on various benchmarks \cite{nigam2022tartarus}. We chose to compare our method to a genetic algorithm, because of their prevalence, their performance \cite{tripp2023genetic} and because, like our method they do not require any training other than that of the proxy model. We ran the algorithm directly with DFT as an evaluation function and with our proxy model; the results are presented in Table~\ref{tab:gap} and in Figure~S\ref{fig:gen_others}. Details of the calculations and JANUS model parameters can be found in the SI. 

As a measure of performance, for each target, we counted the number of molecules within 0.5~eV of the target, the mean absolute distance from the target value and the average Tanimoto distance between Morgan fingerprints of each pair molecules within 0.5~eV of the target. In Table~\ref{tab:gap} we refer to our method as DIDgen, direct inverse design generator. DIDgen and JANUS are both able to significantly increase the proportion of molecules within the target range compared to a random draw of QM9 molecules. Our approach nearly matches or outperforms the genetic algorithm for all 9 metrics in Table~\ref{tab:gap}.

\subsection{logP targeting}

The second task is to target a specific range of octanol-water partition coefficient (logP) values. It is relevant for drug discovery where logP can be used as a measure of cell permeability \cite{wildman1999prediction, lipinski2001lombardo}. Most commercial drugs have a value between 0 and 5. In more recent studies in the field of ML, logP and penalized logP have been used extensively as a benchmark for generative models due to the existence of a cheap empirical model for logP developed by Wildman and Crippen \cite{wildman1999prediction} sometimes called ``Crippen logP'' that is readily available in the RDkit \cite{greg_landrum_2024_10893044}. Here we will use the same target range as \cite{you2018graph} which were used in several recent papers on molecular generation.

We trained "CrippenNet" a GNN developed specifically for this task on a subset of the ZINC dataset\cite{irwin2012zinc, gomez2018automatic} and QM9\cite{ramakrishnan2014quantum}. More details about CrippenNet can be found in the Methods and in the supplementary information. For each of the two target ranges ([-2.5, -2], [5, 5.5]) we generated 1000 molecules and evaluated their diversity using the average pairwise Tanimoto distance--we evaluated the diversity of all generated molecules, not only the ones in the target range to be consistent with ref~\cite{you2018graph}. We limited the molecule size to 85 atoms (including hydrogens), again to be consistent with ref~\cite{you2018graph}. We initialized the generation with random molecules from QM9 because CrippenNet performed significantly better on these molecules. 
\begin{table}[h]
\begin{ruledtabular}
\caption{logP targeting to a certain range. Modified from Ref.~\cite{kong2023molecule}.}
\label{tab:logp}
\begin{tabular}{lccccc}
\textbf{Method} & Evaluation & \multicolumn{2}{c}{$-2.5 \le$ logP $\le -2$} & \multicolumn{2}{c}{$5 \le$ logP $\le 5.5$} \\
& & \multicolumn{1}{c}{Success} & \multicolumn{1}{c}{Diversity} & \multicolumn{1}{c}{Success} & \multicolumn{1}{c}{Diversity}\\
\colrule
ZINC &  & 0.4$\%$ & 0.919 &$1.3\%$ & 0.901 \\
\colrule
ORGAN~\cite{guimaraes2017objective}   & Oracle & 0 & $-$ &$0.2\%$ & 0.909 \\
GCPN~\cite{you2018graph}  & Oracle &$85.5\%$ & $0.392$ &$54.7\%$ & 0.855 \\
SGDS~\cite{kong2023molecule} & Oracle & $\mathbf{86.0\%}$  & 0.874 &$\mathbf{62.2\%}$ & 0.858\\
\colrule
JT-VAE~\cite{jin2018junction} & Proxy & 11.3$\%$ & 0.846 &$7.6\%$ & 0.907 \\
LIMO~\cite{eckmann2022limo} & Proxy & $10.4\%$ & 0.914 &$-$& $-$ \\
DIDgen & Proxy & $\mathbf{43.5\%}$ & \textbf{0.932} & $\mathbf{14.4\%}$ & \textbf{0.917} \\
\end{tabular}
\end{ruledtabular}
\end{table}

The results are presented in Table~\ref{tab:logp} in comparison with refs.~\cite{guimaraes2017objective, you2018graph, kong2023molecule, jin2018junction, eckmann2022limo}. DIDgen generates the most diverse molecules of all methods for both target ranges. When compared to other methods that use a trained proxy model as a predictor for logP, as opposed to the ground truth empirical model which we termed ``oracle'' in Table~\ref{tab:logp}, DIDgen shows the highest performance by a factor of x4 and x2 respectively. However, it does not have a higher success rate then GCPN and SGDS which both use the oracle directly. This is not surprising since the success rate relies heavily on the proxy model performance and generalizability. All methods except JT-VAE could technically use CrippenNet which would offer a way to compare the generation schemes themselves separately of the proxy model's performance (like we did for the energy gap task).

\section{Discussion}

Generative methods like ours that use a learned proxy rely heavily on its performance, especially its ability to generalize in order to hit target properties. Models that can directly use an oracle do not always suffer from this limitation. For example, in the logP targeting task, the oracle was not computationally limiting and thus the number of evaluations was not taken into account. In that case, oracle-based methods had superior success rates. But since evaluations are not costly, as pointed out by other groups of authors \cite{eckmann2022limo}, it would be easy to generate more molecules and filter out molecules that don't meet the criteria. 

In reality, application-relevant properties are often computationally expensive and directly optimizing them is not an option, this is actually the case with the water-octanol partition (logP) which is difficult to evaluate precisely beyond Wildman and Crippen's model \cite{Plante2018jplogp}. The energy gap task is another example of that. In that case, we have seen that JANUS was able to find significantly more molecules in the target range when using a proxy than when using DFT directly. So, even though JANUS does not \textit{require} the use of a proxy, it clearly benefits from its use. This means that in a lot of applications methods that can use an oracle will also end up being dependant on the performance and generalizability of an ML proxy.  

In that same vein, inverting a predictive ML model with our method can be a good way to identify its weaknesses. For example, in the energy gap task, many of the generated molecules for which the proxy model was least accurate contained an O-O bond. The presence of that bond is particularly important for the large gaps (9.3~eV). As illustrated in Figure~\ref{fig:gen}, there is a group of generated molecules containing O-O bonds around a DFT gap of 8~eV which are all predicted to have a gap of around 9.3~eV. In the future, our method could be used as part of an active learning loop to iteratively improve the model generalizability, in this case, by retraining on a set of molecules containing O-O bonds, for example.

Another way to evaluate the performance of generative models that is independent of the evaluation method is to measure their ability to generate diverse molecules. On that metric, our method performed better than all other methods we have tested it against. This is important because for any application, new materials must simultaneously meet multiple criteria that can't all be modeled and optimized for. For example, molecules for display applications must emit light of a certain color, but they must also be synthesizable, bright, stable, cheap, safe, soluble, etc. Generating a more diverse set of molecules simply increases the chances that one of the generated molecules will meet multiple of these criteria.

The inversion procedure presented here can be used on any GNN architecture that uses molecular 2D graphs (like SMILES), but it requires modifying them so that they use an explicit representation of the adjacency matrix which may require retraining. This can be a limiting factor for the adoption of our procedure. To give an example of how this process can be done we make available on our public repository a version of the popular CGCNN model \cite{xie2018crystal} using an explicit adjacency matrix.

%
In conclusion, we demonstrated that by carefully constraining the representation of molecules, a property predicting GNN can be turned into a diverse conditional generator without any additional training. Our method is a literal implementation of the inverse design paradigm: properties are put in and materials are obtained going backwards through a pipeline that is built to do the opposite. Although it is a ``naive'' way to approach this problem, we showed that it rivals with complex state-of-the-art models. Ultimately, we hope that will help accelerate the design and discovery of functional materials.  

\section{Methods} \label{seq:methods}

\subsection{Training of the property predictors}
Since optimizing the inputs (generating molecules) will require an explicit representation of the adjacency matrix, sparse matrices and lists of edge indices cannot be used during training of the property predictor GNN. Mini-batching using a single adjacency matrix per batch thus uses a large amount of memory that scales quadratically with the total number of atoms in a batch. To avoid that issue we use a fixed graph size ($N$) instead which allows us to store the adjacency matrices as a $B \times N \times N$ array where B is the batch size. Empty rows of the adjacency matrix are associated with atoms of type "no-atom" in the feature vector one-hot encoding. This allows for molecules of any size smaller than $N$. 

Although sloped rounding and maximum functions guarantee that the adjacency matrix and the feature vector are populated with near-integer values, these values are not exactly integers and these variations around integer values can have a significant impact on the property predicated by the GNN. To make the GNN less sensitive to these variations, we add a small amount of random noise around integer values of $\mathbb{A}$ and $\mathbb{F}$ during training.

For the energy gap task, we used a simple graph convolutional network (GCN) with an adjustable number of convolutions, a pooling layer and an adjustable number of linear layers. We trained this model on the QM9 dataset containing approximately 130000 small molecules for which the energy gap was calculated with density functional theory (DFT). With manual hyperparameter tuning, we obtain a mean absolute error of about 0.12~eV on the test set which is on par with other models using the same amount of information (2D graphs, no coordinates) \cite{ramakrishnan2014quantum}. More details on this model can be found in the SI.

For the LogP task, we use CrippenNet a graph neural network that we developed based on the structure of Crippen's empirical model. CrippenNet classifies atoms inside molecules as one of 69 classes associated with a logP value and then sums the individual contributions of each atom to obtain the total logP of the molecules. In the empirical model, each of these classes is defined by certain properties described by SMARTS string. For example, class \#18 is an aromatic carbon atom and its associated logP is 0.1581 \cite{wildman1999prediction}. In CrippenNet the classes are learned from the graph of the molecule by training on a subset of the ZINC dataset \cite{irwin2012zinc} containing 250,000 molecules (the same subset used in Ref.~\cite{gomez2018automatic}) and QM9. More details on CrippenNet can be found in the supplemental information. Our goal was to get a model that resembles the ground truth model as much as possible in order to be as close as possible to directly inverting it. Note that it may be possible to obtain a perfect classification from the graph representation of the molecule in which case DIDgen's success rate would be 100\%. This would require a careful analysis of the SMARTS classes and their resulting convolutions.

\subsection{DFT validation of the energy gap}

For each generated molecule, we obtained the conformation (3D atomic positions) using the RDKit. In some cases no conformation could be obtained from the generated molecule. In those cases, the molecule was discarded and a new molecule was generated until a total of 100 molecules with conformation were obtained for each target.

For all 300 molecules, we performed density functional theory calculations to obtain the "true" HOMO-LUMO gap. For consistency, we used DFT parameters as close as possible to Ramakrishnan et al. \cite{ramakrishnan2014quantum}. We fully relaxed the molecules and obtained the HOMO-LUMO gap with B3LYP/6-31G(2df,2p) as implemented in ORCA \cite{neese2020orca}.

\subsection{Loss function for the inversion procedure}

Since chemically valid molecules are not always realistic (they don't necessarily have a stable conformation), in order to guide the generation towards more realistic molecules, we added an additional term to the loss function. We used the average composition, $C_i$, in QM9 for each element $i$ as an objective. The total loss for this task was the following:
\begin{equation}
\mathbb{L} = \frac{\lambda_l \mathbb{L}_l^2 + \lambda_s \mathbb{L}_s^2 + \lambda_c \mathbb{L}_c^2}{\lambda_l + \lambda_s + \lambda_c},
\end{equation}
where
\begin{equation}
\mathbb{L}_c = \sum_i \frac{\sum_j \mathbb{F}_{ij}}{\sum_{ij} \mathbb{F}_{ij}} - C_i,
\end{equation}
$\mathbb{L}_l$ is the root mean square error with respect to the target p, $\mathbb{L}_s$ is the loss associated with the maximum valence defined in equation~\ref{eq:constraint} of the supplementary information and $\lambda_{l,s,c}$ are adjustable hyperparameters.

\section{Code Availability}

The code used for the experiments presented in this paper is available at \href{https://github.com/ftherrien/inv-design}{github.com/ftherrien/inv-design}.

\section{Acknowledgements}

We thank Dr. Alexander Davis from the University of Toronto for insightful discussions, ideas and valuable feedback. The authors acknowledge support from The Alliance for AI-Accelerated Materials Discovery (A3MD), which includes funding from Total Energies SE., BP, Meta, and LG AI Research.

\bibliography{references}

\begin{thebibliography}{51}%
\makeatletter
\providecommand \@ifxundefined [1]{%
 \@ifx{#1\undefined}
}%
\providecommand \@ifnum [1]{%
 \ifnum #1\expandafter \@firstoftwo
 \else \expandafter \@secondoftwo
 \fi
}%
\providecommand \@ifx [1]{%
 \ifx #1\expandafter \@firstoftwo
 \else \expandafter \@secondoftwo
 \fi
}%
\providecommand \natexlab [1]{#1}%
\providecommand \enquote  [1]{``#1''}%
\providecommand \bibnamefont  [1]{#1}%
\providecommand \bibfnamefont [1]{#1}%
\providecommand \citenamefont [1]{#1}%
\providecommand \href@noop [0]{\@secondoftwo}%
\providecommand \href [0]{\begingroup \@sanitize@url \@href}%
\providecommand \@href[1]{\@@startlink{#1}\@@href}%
\providecommand \@@href[1]{\endgroup#1\@@endlink}%
\providecommand \@sanitize@url [0]{\catcode `\\12\catcode `\$12\catcode
  `\&12\catcode `\#12\catcode `\^12\catcode `\_12\catcode `\%12\relax}%
\providecommand \@@startlink[1]{}%
\providecommand \@@endlink[0]{}%
\providecommand \url  [0]{\begingroup\@sanitize@url \@url }%
\providecommand \@url [1]{\endgroup\@href {#1}{\urlprefix }}%
\providecommand \urlprefix  [0]{URL }%
\providecommand \Eprint [0]{\href }%
\providecommand \doibase [0]{https://doi.org/}%
\providecommand \selectlanguage [0]{\@gobble}%
\providecommand \bibinfo  [0]{\@secondoftwo}%
\providecommand \bibfield  [0]{\@secondoftwo}%
\providecommand \translation [1]{[#1]}%
\providecommand \BibitemOpen [0]{}%
\providecommand \bibitemStop [0]{}%
\providecommand \bibitemNoStop [0]{.\EOS\space}%
\providecommand \EOS [0]{\spacefactor3000\relax}%
\providecommand \BibitemShut  [1]{\csname bibitem#1\endcsname}%
\let\auto@bib@innerbib\@empty
\bibitem [{\citenamefont {Zunger}(2018)}]{zunger2018inverse}%
  \BibitemOpen
  \bibfield  {author} {\bibinfo {author} {\bibfnamefont {A.}~\bibnamefont
  {Zunger}},\ }\bibfield  {title} {\bibinfo {title} {Inverse design in search
  of materials with target functionalities},\ }\href@noop {} {\bibfield
  {journal} {\bibinfo  {journal} {Nature Reviews Chemistry}\ }\textbf {\bibinfo
  {volume} {2}},\ \bibinfo {pages} {1} (\bibinfo {year} {2018})}\BibitemShut
  {NoStop}%
\bibitem [{\citenamefont {Lu}(2021)}]{lu2021computational}%
  \BibitemOpen
  \bibfield  {author} {\bibinfo {author} {\bibfnamefont {Z.}~\bibnamefont
  {Lu}},\ }\bibfield  {title} {\bibinfo {title} {Computational discovery of
  energy materials in the era of big data and machine learning: a critical
  review},\ }\href@noop {} {\bibfield  {journal} {\bibinfo  {journal}
  {Materials Reports: Energy}\ }\textbf {\bibinfo {volume} {1}},\ \bibinfo
  {pages} {100047} (\bibinfo {year} {2021})}\BibitemShut {NoStop}%
\bibitem [{\citenamefont {Therrien}\ \emph {et~al.}(2021)\citenamefont
  {Therrien}, \citenamefont {Jones},\ and\ \citenamefont
  {Stevanovi{\'c}}}]{therrien2021metastable}%
  \BibitemOpen
  \bibfield  {author} {\bibinfo {author} {\bibfnamefont {F.}~\bibnamefont
  {Therrien}}, \bibinfo {author} {\bibfnamefont {E.~B.}\ \bibnamefont
  {Jones}},\ and\ \bibinfo {author} {\bibfnamefont {V.}~\bibnamefont
  {Stevanovi{\'c}}},\ }\bibfield  {title} {\bibinfo {title} {Metastable
  materials discovery in the age of large-scale computation},\ }\href@noop {}
  {\bibfield  {journal} {\bibinfo  {journal} {Applied Physics Reviews}\
  }\textbf {\bibinfo {volume} {8}} (\bibinfo {year} {2021})}\BibitemShut
  {NoStop}%
\bibitem [{\citenamefont {Gorai}\ \emph {et~al.}(2017)\citenamefont {Gorai},
  \citenamefont {Stevanovi{\'c}},\ and\ \citenamefont
  {Toberer}}]{gorai2017computationally}%
  \BibitemOpen
  \bibfield  {author} {\bibinfo {author} {\bibfnamefont {P.}~\bibnamefont
  {Gorai}}, \bibinfo {author} {\bibfnamefont {V.}~\bibnamefont
  {Stevanovi{\'c}}},\ and\ \bibinfo {author} {\bibfnamefont {E.~S.}\
  \bibnamefont {Toberer}},\ }\bibfield  {title} {\bibinfo {title}
  {Computationally guided discovery of thermoelectric materials},\ }\href@noop
  {} {\bibfield  {journal} {\bibinfo  {journal} {Nature Reviews Materials}\
  }\textbf {\bibinfo {volume} {2}},\ \bibinfo {pages} {1} (\bibinfo {year}
  {2017})}\BibitemShut {NoStop}%
\bibitem [{\citenamefont {Garrity}(2018)}]{garrity2018high}%
  \BibitemOpen
  \bibfield  {author} {\bibinfo {author} {\bibfnamefont {K.~F.}\ \bibnamefont
  {Garrity}},\ }\bibfield  {title} {\bibinfo {title} {High-throughput
  first-principles search for new ferroelectrics},\ }\href@noop {} {\bibfield
  {journal} {\bibinfo  {journal} {Physical Review B}\ }\textbf {\bibinfo
  {volume} {97}},\ \bibinfo {pages} {024115} (\bibinfo {year}
  {2018})}\BibitemShut {NoStop}%
\bibitem [{\citenamefont {Curtarolo}\ \emph {et~al.}(2013)\citenamefont
  {Curtarolo}, \citenamefont {Hart}, \citenamefont {Nardelli}, \citenamefont
  {Mingo}, \citenamefont {Sanvito},\ and\ \citenamefont
  {Levy}}]{curtarolo2013high}%
  \BibitemOpen
  \bibfield  {author} {\bibinfo {author} {\bibfnamefont {S.}~\bibnamefont
  {Curtarolo}}, \bibinfo {author} {\bibfnamefont {G.~L.}\ \bibnamefont {Hart}},
  \bibinfo {author} {\bibfnamefont {M.~B.}\ \bibnamefont {Nardelli}}, \bibinfo
  {author} {\bibfnamefont {N.}~\bibnamefont {Mingo}}, \bibinfo {author}
  {\bibfnamefont {S.}~\bibnamefont {Sanvito}},\ and\ \bibinfo {author}
  {\bibfnamefont {O.}~\bibnamefont {Levy}},\ }\bibfield  {title} {\bibinfo
  {title} {The high-throughput highway to computational materials design},\
  }\href@noop {} {\bibfield  {journal} {\bibinfo  {journal} {Nature materials}\
  }\textbf {\bibinfo {volume} {12}},\ \bibinfo {pages} {191} (\bibinfo {year}
  {2013})}\BibitemShut {NoStop}%
\bibitem [{\citenamefont {Alberi}\ \emph {et~al.}(2018)\citenamefont {Alberi},
  \citenamefont {Nardelli}, \citenamefont {Zakutayev}, \citenamefont {Mitas},
  \citenamefont {Curtarolo}, \citenamefont {Jain}, \citenamefont {Fornari},
  \citenamefont {Marzari}, \citenamefont {Takeuchi}, \citenamefont {Green}
  \emph {et~al.}}]{alberi20182019}%
  \BibitemOpen
  \bibfield  {author} {\bibinfo {author} {\bibfnamefont {K.}~\bibnamefont
  {Alberi}}, \bibinfo {author} {\bibfnamefont {M.~B.}\ \bibnamefont
  {Nardelli}}, \bibinfo {author} {\bibfnamefont {A.}~\bibnamefont {Zakutayev}},
  \bibinfo {author} {\bibfnamefont {L.}~\bibnamefont {Mitas}}, \bibinfo
  {author} {\bibfnamefont {S.}~\bibnamefont {Curtarolo}}, \bibinfo {author}
  {\bibfnamefont {A.}~\bibnamefont {Jain}}, \bibinfo {author} {\bibfnamefont
  {M.}~\bibnamefont {Fornari}}, \bibinfo {author} {\bibfnamefont
  {N.}~\bibnamefont {Marzari}}, \bibinfo {author} {\bibfnamefont
  {I.}~\bibnamefont {Takeuchi}}, \bibinfo {author} {\bibfnamefont {M.~L.}\
  \bibnamefont {Green}}, \emph {et~al.},\ }\bibfield  {title} {\bibinfo {title}
  {The 2019 materials by design roadmap},\ }\href@noop {} {\bibfield  {journal}
  {\bibinfo  {journal} {Journal of Physics D: Applied Physics}\ }\textbf
  {\bibinfo {volume} {52}},\ \bibinfo {pages} {013001} (\bibinfo {year}
  {2018})}\BibitemShut {NoStop}%
\bibitem [{\citenamefont {Schmidt}\ \emph {et~al.}(2019)\citenamefont
  {Schmidt}, \citenamefont {Marques}, \citenamefont {Botti},\ and\
  \citenamefont {Marques}}]{schmidt2019recent}%
  \BibitemOpen
  \bibfield  {author} {\bibinfo {author} {\bibfnamefont {J.}~\bibnamefont
  {Schmidt}}, \bibinfo {author} {\bibfnamefont {M.~R.}\ \bibnamefont
  {Marques}}, \bibinfo {author} {\bibfnamefont {S.}~\bibnamefont {Botti}},\
  and\ \bibinfo {author} {\bibfnamefont {M.~A.}\ \bibnamefont {Marques}},\
  }\bibfield  {title} {\bibinfo {title} {Recent advances and applications of
  machine learning in solid-state materials science},\ }\href@noop {}
  {\bibfield  {journal} {\bibinfo  {journal} {npj Computational Materials}\
  }\textbf {\bibinfo {volume} {5}},\ \bibinfo {pages} {1} (\bibinfo {year}
  {2019})}\BibitemShut {NoStop}%
\bibitem [{\citenamefont {Butler}\ \emph {et~al.}(2018)\citenamefont {Butler},
  \citenamefont {Davies}, \citenamefont {Cartwright}, \citenamefont {Isayev},\
  and\ \citenamefont {Walsh}}]{butler2018machine}%
  \BibitemOpen
  \bibfield  {author} {\bibinfo {author} {\bibfnamefont {K.~T.}\ \bibnamefont
  {Butler}}, \bibinfo {author} {\bibfnamefont {D.~W.}\ \bibnamefont {Davies}},
  \bibinfo {author} {\bibfnamefont {H.}~\bibnamefont {Cartwright}}, \bibinfo
  {author} {\bibfnamefont {O.}~\bibnamefont {Isayev}},\ and\ \bibinfo {author}
  {\bibfnamefont {A.}~\bibnamefont {Walsh}},\ }\bibfield  {title} {\bibinfo
  {title} {Machine learning for molecular and materials science},\ }\href@noop
  {} {\bibfield  {journal} {\bibinfo  {journal} {Nature}\ }\textbf {\bibinfo
  {volume} {559}},\ \bibinfo {pages} {547} (\bibinfo {year}
  {2018})}\BibitemShut {NoStop}%
\bibitem [{\citenamefont {Reiser}\ \emph {et~al.}(2022)\citenamefont {Reiser},
  \citenamefont {Neubert}, \citenamefont {Eberhard}, \citenamefont {Torresi},
  \citenamefont {Zhou}, \citenamefont {Shao}, \citenamefont {Metni},
  \citenamefont {van Hoesel}, \citenamefont {Schopmans}, \citenamefont {Sommer}
  \emph {et~al.}}]{reiser2022graph}%
  \BibitemOpen
  \bibfield  {author} {\bibinfo {author} {\bibfnamefont {P.}~\bibnamefont
  {Reiser}}, \bibinfo {author} {\bibfnamefont {M.}~\bibnamefont {Neubert}},
  \bibinfo {author} {\bibfnamefont {A.}~\bibnamefont {Eberhard}}, \bibinfo
  {author} {\bibfnamefont {L.}~\bibnamefont {Torresi}}, \bibinfo {author}
  {\bibfnamefont {C.}~\bibnamefont {Zhou}}, \bibinfo {author} {\bibfnamefont
  {C.}~\bibnamefont {Shao}}, \bibinfo {author} {\bibfnamefont {H.}~\bibnamefont
  {Metni}}, \bibinfo {author} {\bibfnamefont {C.}~\bibnamefont {van Hoesel}},
  \bibinfo {author} {\bibfnamefont {H.}~\bibnamefont {Schopmans}}, \bibinfo
  {author} {\bibfnamefont {T.}~\bibnamefont {Sommer}}, \emph {et~al.},\
  }\bibfield  {title} {\bibinfo {title} {Graph neural networks for materials
  science and chemistry},\ }\href@noop {} {\bibfield  {journal} {\bibinfo
  {journal} {Communications Materials}\ }\textbf {\bibinfo {volume} {3}},\
  \bibinfo {pages} {93} (\bibinfo {year} {2022})}\BibitemShut {NoStop}%
\bibitem [{\citenamefont {Wigh}\ \emph {et~al.}(2022)\citenamefont {Wigh},
  \citenamefont {Goodman},\ and\ \citenamefont {Lapkin}}]{wigh2022review}%
  \BibitemOpen
  \bibfield  {author} {\bibinfo {author} {\bibfnamefont {D.~S.}\ \bibnamefont
  {Wigh}}, \bibinfo {author} {\bibfnamefont {J.~M.}\ \bibnamefont {Goodman}},\
  and\ \bibinfo {author} {\bibfnamefont {A.~A.}\ \bibnamefont {Lapkin}},\
  }\bibfield  {title} {\bibinfo {title} {A review of molecular representation
  in the age of machine learning},\ }\href@noop {} {\bibfield  {journal}
  {\bibinfo  {journal} {Wiley Interdisciplinary Reviews: Computational
  Molecular Science}\ }\textbf {\bibinfo {volume} {12}},\ \bibinfo {pages}
  {e1603} (\bibinfo {year} {2022})}\BibitemShut {NoStop}%
\bibitem [{\citenamefont {Wei}\ \emph {et~al.}(2019)\citenamefont {Wei},
  \citenamefont {Chu}, \citenamefont {Sun}, \citenamefont {Xu}, \citenamefont
  {Deng}, \citenamefont {Chen}, \citenamefont {Wei},\ and\ \citenamefont
  {Lei}}]{wei2019machine}%
  \BibitemOpen
  \bibfield  {author} {\bibinfo {author} {\bibfnamefont {J.}~\bibnamefont
  {Wei}}, \bibinfo {author} {\bibfnamefont {X.}~\bibnamefont {Chu}}, \bibinfo
  {author} {\bibfnamefont {X.-Y.}\ \bibnamefont {Sun}}, \bibinfo {author}
  {\bibfnamefont {K.}~\bibnamefont {Xu}}, \bibinfo {author} {\bibfnamefont
  {H.-X.}\ \bibnamefont {Deng}}, \bibinfo {author} {\bibfnamefont
  {J.}~\bibnamefont {Chen}}, \bibinfo {author} {\bibfnamefont {Z.}~\bibnamefont
  {Wei}},\ and\ \bibinfo {author} {\bibfnamefont {M.}~\bibnamefont {Lei}},\
  }\bibfield  {title} {\bibinfo {title} {Machine learning in materials
  science},\ }\href@noop {} {\bibfield  {journal} {\bibinfo  {journal}
  {InfoMat}\ }\textbf {\bibinfo {volume} {1}},\ \bibinfo {pages} {338}
  (\bibinfo {year} {2019})}\BibitemShut {NoStop}%
\bibitem [{\citenamefont {Paszke}\ \emph {et~al.}(2019)\citenamefont {Paszke},
  \citenamefont {Gross}, \citenamefont {Massa}, \citenamefont {Lerer},
  \citenamefont {Bradbury}, \citenamefont {Chanan}, \citenamefont {Killeen},
  \citenamefont {Lin}, \citenamefont {Gimelshein}, \citenamefont {Antiga} \emph
  {et~al.}}]{paszke2019pytorch}%
  \BibitemOpen
  \bibfield  {author} {\bibinfo {author} {\bibfnamefont {A.}~\bibnamefont
  {Paszke}}, \bibinfo {author} {\bibfnamefont {S.}~\bibnamefont {Gross}},
  \bibinfo {author} {\bibfnamefont {F.}~\bibnamefont {Massa}}, \bibinfo
  {author} {\bibfnamefont {A.}~\bibnamefont {Lerer}}, \bibinfo {author}
  {\bibfnamefont {J.}~\bibnamefont {Bradbury}}, \bibinfo {author}
  {\bibfnamefont {G.}~\bibnamefont {Chanan}}, \bibinfo {author} {\bibfnamefont
  {T.}~\bibnamefont {Killeen}}, \bibinfo {author} {\bibfnamefont
  {Z.}~\bibnamefont {Lin}}, \bibinfo {author} {\bibfnamefont {N.}~\bibnamefont
  {Gimelshein}}, \bibinfo {author} {\bibfnamefont {L.}~\bibnamefont {Antiga}},
  \emph {et~al.},\ }\bibfield  {title} {\bibinfo {title} {Pytorch: An
  imperative style, high-performance deep learning library},\ }\href@noop {}
  {\bibfield  {journal} {\bibinfo  {journal} {Advances in neural information
  processing systems}\ }\textbf {\bibinfo {volume} {32}} (\bibinfo {year}
  {2019})}\BibitemShut {NoStop}%
\bibitem [{\citenamefont {Abadi}\ \emph {et~al.}(2015)\citenamefont {Abadi},
  \citenamefont {Agarwal}, \citenamefont {Barham}, \citenamefont {Brevdo},
  \citenamefont {Chen}, \citenamefont {Citro}, \citenamefont {Corrado},
  \citenamefont {Davis}, \citenamefont {Dean}, \citenamefont {Devin},
  \citenamefont {Ghemawat}, \citenamefont {Goodfellow}, \citenamefont {Harp},
  \citenamefont {Irving}, \citenamefont {Isard}, \citenamefont {Jia},
  \citenamefont {Jozefowicz}, \citenamefont {Kaiser}, \citenamefont {Kudlur},
  \citenamefont {Levenberg}, \citenamefont {Man\'{e}}, \citenamefont {Monga},
  \citenamefont {Moore}, \citenamefont {Murray}, \citenamefont {Olah},
  \citenamefont {Schuster}, \citenamefont {Shlens}, \citenamefont {Steiner},
  \citenamefont {Sutskever}, \citenamefont {Talwar}, \citenamefont {Tucker},
  \citenamefont {Vanhoucke}, \citenamefont {Vasudevan}, \citenamefont
  {Vi\'{e}gas}, \citenamefont {Vinyals}, \citenamefont {Warden}, \citenamefont
  {Wattenberg}, \citenamefont {Wicke}, \citenamefont {Yu},\ and\ \citenamefont
  {Zheng}}]{tensorflow2015-whitepaper}%
  \BibitemOpen
  \bibfield  {author} {\bibinfo {author} {\bibfnamefont {M.}~\bibnamefont
  {Abadi}}, \bibinfo {author} {\bibfnamefont {A.}~\bibnamefont {Agarwal}},
  \bibinfo {author} {\bibfnamefont {P.}~\bibnamefont {Barham}}, \bibinfo
  {author} {\bibfnamefont {E.}~\bibnamefont {Brevdo}}, \bibinfo {author}
  {\bibfnamefont {Z.}~\bibnamefont {Chen}}, \bibinfo {author} {\bibfnamefont
  {C.}~\bibnamefont {Citro}}, \bibinfo {author} {\bibfnamefont {G.~S.}\
  \bibnamefont {Corrado}}, \bibinfo {author} {\bibfnamefont {A.}~\bibnamefont
  {Davis}}, \bibinfo {author} {\bibfnamefont {J.}~\bibnamefont {Dean}},
  \bibinfo {author} {\bibfnamefont {M.}~\bibnamefont {Devin}}, \bibinfo
  {author} {\bibfnamefont {S.}~\bibnamefont {Ghemawat}}, \bibinfo {author}
  {\bibfnamefont {I.}~\bibnamefont {Goodfellow}}, \bibinfo {author}
  {\bibfnamefont {A.}~\bibnamefont {Harp}}, \bibinfo {author} {\bibfnamefont
  {G.}~\bibnamefont {Irving}}, \bibinfo {author} {\bibfnamefont
  {M.}~\bibnamefont {Isard}}, \bibinfo {author} {\bibfnamefont
  {Y.}~\bibnamefont {Jia}}, \bibinfo {author} {\bibfnamefont {R.}~\bibnamefont
  {Jozefowicz}}, \bibinfo {author} {\bibfnamefont {L.}~\bibnamefont {Kaiser}},
  \bibinfo {author} {\bibfnamefont {M.}~\bibnamefont {Kudlur}}, \bibinfo
  {author} {\bibfnamefont {J.}~\bibnamefont {Levenberg}}, \bibinfo {author}
  {\bibfnamefont {D.}~\bibnamefont {Man\'{e}}}, \bibinfo {author}
  {\bibfnamefont {R.}~\bibnamefont {Monga}}, \bibinfo {author} {\bibfnamefont
  {S.}~\bibnamefont {Moore}}, \bibinfo {author} {\bibfnamefont
  {D.}~\bibnamefont {Murray}}, \bibinfo {author} {\bibfnamefont
  {C.}~\bibnamefont {Olah}}, \bibinfo {author} {\bibfnamefont {M.}~\bibnamefont
  {Schuster}}, \bibinfo {author} {\bibfnamefont {J.}~\bibnamefont {Shlens}},
  \bibinfo {author} {\bibfnamefont {B.}~\bibnamefont {Steiner}}, \bibinfo
  {author} {\bibfnamefont {I.}~\bibnamefont {Sutskever}}, \bibinfo {author}
  {\bibfnamefont {K.}~\bibnamefont {Talwar}}, \bibinfo {author} {\bibfnamefont
  {P.}~\bibnamefont {Tucker}}, \bibinfo {author} {\bibfnamefont
  {V.}~\bibnamefont {Vanhoucke}}, \bibinfo {author} {\bibfnamefont
  {V.}~\bibnamefont {Vasudevan}}, \bibinfo {author} {\bibfnamefont
  {F.}~\bibnamefont {Vi\'{e}gas}}, \bibinfo {author} {\bibfnamefont
  {O.}~\bibnamefont {Vinyals}}, \bibinfo {author} {\bibfnamefont
  {P.}~\bibnamefont {Warden}}, \bibinfo {author} {\bibfnamefont
  {M.}~\bibnamefont {Wattenberg}}, \bibinfo {author} {\bibfnamefont
  {M.}~\bibnamefont {Wicke}}, \bibinfo {author} {\bibfnamefont
  {Y.}~\bibnamefont {Yu}},\ and\ \bibinfo {author} {\bibfnamefont
  {X.}~\bibnamefont {Zheng}},\ }\href {https://www.tensorflow.org/} {\bibinfo
  {title} {{TensorFlow}: Large-scale machine learning on heterogeneous
  systems}} (\bibinfo {year} {2015}),\ \bibinfo {note} {software available from
  tensorflow.org}\BibitemShut {NoStop}%
\bibitem [{\citenamefont {pandas~development team}(2020)}]{reback2020pandas}%
  \BibitemOpen
  \bibfield  {author} {\bibinfo {author} {\bibfnamefont {T.}~\bibnamefont
  {pandas~development team}},\ }\href {https://doi.org/10.5281/zenodo.3509134}
  {\bibinfo {title} {pandas-dev/pandas: Pandas}} (\bibinfo {year}
  {2020})\BibitemShut {NoStop}%
\bibitem [{\citenamefont {Li}\ \emph {et~al.}(2023)\citenamefont {Li},
  \citenamefont {DeCost}, \citenamefont {Choudhary}, \citenamefont
  {Greenwood},\ and\ \citenamefont {Hattrick-Simpers}}]{li2023critical}%
  \BibitemOpen
  \bibfield  {author} {\bibinfo {author} {\bibfnamefont {K.}~\bibnamefont
  {Li}}, \bibinfo {author} {\bibfnamefont {B.}~\bibnamefont {DeCost}}, \bibinfo
  {author} {\bibfnamefont {K.}~\bibnamefont {Choudhary}}, \bibinfo {author}
  {\bibfnamefont {M.}~\bibnamefont {Greenwood}},\ and\ \bibinfo {author}
  {\bibfnamefont {J.}~\bibnamefont {Hattrick-Simpers}},\ }\bibfield  {title}
  {\bibinfo {title} {A critical examination of robustness and generalizability
  of machine learning prediction of materials properties},\ }\href@noop {}
  {\bibfield  {journal} {\bibinfo  {journal} {npj Computational Materials}\
  }\textbf {\bibinfo {volume} {9}},\ \bibinfo {pages} {55} (\bibinfo {year}
  {2023})}\BibitemShut {NoStop}%
\bibitem [{\citenamefont {Goedecker}(2004)}]{goedecker2004minima}%
  \BibitemOpen
  \bibfield  {author} {\bibinfo {author} {\bibfnamefont {S.}~\bibnamefont
  {Goedecker}},\ }\bibfield  {title} {\bibinfo {title} {Minima hopping: An
  efficient search method for the global minimum of the potential energy
  surface of complex molecular systems},\ }\href@noop {} {\bibfield  {journal}
  {\bibinfo  {journal} {The Journal of chemical physics}\ }\textbf {\bibinfo
  {volume} {120}},\ \bibinfo {pages} {9911} (\bibinfo {year}
  {2004})}\BibitemShut {NoStop}%
\bibitem [{\citenamefont {Marto{\v{n}}{\'a}k}\ \emph
  {et~al.}(2003)\citenamefont {Marto{\v{n}}{\'a}k}, \citenamefont {Laio},\ and\
  \citenamefont {Parrinello}}]{martovnak2003predicting}%
  \BibitemOpen
  \bibfield  {author} {\bibinfo {author} {\bibfnamefont {R.}~\bibnamefont
  {Marto{\v{n}}{\'a}k}}, \bibinfo {author} {\bibfnamefont {A.}~\bibnamefont
  {Laio}},\ and\ \bibinfo {author} {\bibfnamefont {M.}~\bibnamefont
  {Parrinello}},\ }\bibfield  {title} {\bibinfo {title} {Predicting crystal
  structures: the parrinello-rahman method revisited},\ }\href@noop {}
  {\bibfield  {journal} {\bibinfo  {journal} {Physical review letters}\
  }\textbf {\bibinfo {volume} {90}},\ \bibinfo {pages} {075503} (\bibinfo
  {year} {2003})}\BibitemShut {NoStop}%
\bibitem [{\citenamefont {Glass}\ \emph {et~al.}(2006)\citenamefont {Glass},
  \citenamefont {Oganov},\ and\ \citenamefont {Hansen}}]{glass2006uspex}%
  \BibitemOpen
  \bibfield  {author} {\bibinfo {author} {\bibfnamefont {C.~W.}\ \bibnamefont
  {Glass}}, \bibinfo {author} {\bibfnamefont {A.~R.}\ \bibnamefont {Oganov}},\
  and\ \bibinfo {author} {\bibfnamefont {N.}~\bibnamefont {Hansen}},\
  }\bibfield  {title} {\bibinfo {title} {Uspex—evolutionary crystal structure
  prediction},\ }\href@noop {} {\bibfield  {journal} {\bibinfo  {journal}
  {Computer physics communications}\ }\textbf {\bibinfo {volume} {175}},\
  \bibinfo {pages} {713} (\bibinfo {year} {2006})}\BibitemShut {NoStop}%
\bibitem [{\citenamefont {Jensen}(2019)}]{jensen2019graph}%
  \BibitemOpen
  \bibfield  {author} {\bibinfo {author} {\bibfnamefont {J.~H.}\ \bibnamefont
  {Jensen}},\ }\bibfield  {title} {\bibinfo {title} {A graph-based genetic
  algorithm and generative model/monte carlo tree search for the exploration of
  chemical space},\ }\href@noop {} {\bibfield  {journal} {\bibinfo  {journal}
  {Chemical science}\ }\textbf {\bibinfo {volume} {10}},\ \bibinfo {pages}
  {3567} (\bibinfo {year} {2019})}\BibitemShut {NoStop}%
\bibitem [{\citenamefont {Oganov}\ \emph {et~al.}(2019)\citenamefont {Oganov},
  \citenamefont {Pickard}, \citenamefont {Zhu},\ and\ \citenamefont
  {Needs}}]{oganov2019structure}%
  \BibitemOpen
  \bibfield  {author} {\bibinfo {author} {\bibfnamefont {A.~R.}\ \bibnamefont
  {Oganov}}, \bibinfo {author} {\bibfnamefont {C.~J.}\ \bibnamefont {Pickard}},
  \bibinfo {author} {\bibfnamefont {Q.}~\bibnamefont {Zhu}},\ and\ \bibinfo
  {author} {\bibfnamefont {R.~J.}\ \bibnamefont {Needs}},\ }\bibfield  {title}
  {\bibinfo {title} {Structure prediction drives materials discovery},\
  }\href@noop {} {\bibfield  {journal} {\bibinfo  {journal} {Nature Reviews
  Materials}\ }\textbf {\bibinfo {volume} {4}},\ \bibinfo {pages} {331}
  (\bibinfo {year} {2019})}\BibitemShut {NoStop}%
\bibitem [{\citenamefont {Sanchez-Lengeling}\ and\ \citenamefont
  {Aspuru-Guzik}(2018)}]{sanchez2018inverse}%
  \BibitemOpen
  \bibfield  {author} {\bibinfo {author} {\bibfnamefont {B.}~\bibnamefont
  {Sanchez-Lengeling}}\ and\ \bibinfo {author} {\bibfnamefont {A.}~\bibnamefont
  {Aspuru-Guzik}},\ }\bibfield  {title} {\bibinfo {title} {Inverse molecular
  design using machine learning: Generative models for matter engineering},\
  }\href@noop {} {\bibfield  {journal} {\bibinfo  {journal} {Science}\ }\textbf
  {\bibinfo {volume} {361}},\ \bibinfo {pages} {360} (\bibinfo {year}
  {2018})}\BibitemShut {NoStop}%
\bibitem [{\citenamefont {Jin}\ \emph {et~al.}(2018)\citenamefont {Jin},
  \citenamefont {Barzilay},\ and\ \citenamefont {Jaakkola}}]{jin2018junction}%
  \BibitemOpen
  \bibfield  {author} {\bibinfo {author} {\bibfnamefont {W.}~\bibnamefont
  {Jin}}, \bibinfo {author} {\bibfnamefont {R.}~\bibnamefont {Barzilay}},\ and\
  \bibinfo {author} {\bibfnamefont {T.}~\bibnamefont {Jaakkola}},\ }\bibfield
  {title} {\bibinfo {title} {Junction tree variational autoencoder for
  molecular graph generation},\ }in\ \href@noop {} {\emph {\bibinfo {booktitle}
  {International conference on machine learning}}}\ (\bibinfo {organization}
  {PMLR},\ \bibinfo {year} {2018})\ pp.\ \bibinfo {pages}
  {2323--2332}\BibitemShut {NoStop}%
\bibitem [{\citenamefont {G{\'o}mez-Bombarelli}\ \emph
  {et~al.}(2018)\citenamefont {G{\'o}mez-Bombarelli}, \citenamefont {Wei},
  \citenamefont {Duvenaud}, \citenamefont {Hern{\'a}ndez-Lobato}, \citenamefont
  {S{\'a}nchez-Lengeling}, \citenamefont {Sheberla}, \citenamefont
  {Aguilera-Iparraguirre}, \citenamefont {Hirzel}, \citenamefont {Adams},\ and\
  \citenamefont {Aspuru-Guzik}}]{gomez2018automatic}%
  \BibitemOpen
  \bibfield  {author} {\bibinfo {author} {\bibfnamefont {R.}~\bibnamefont
  {G{\'o}mez-Bombarelli}}, \bibinfo {author} {\bibfnamefont {J.~N.}\
  \bibnamefont {Wei}}, \bibinfo {author} {\bibfnamefont {D.}~\bibnamefont
  {Duvenaud}}, \bibinfo {author} {\bibfnamefont {J.~M.}\ \bibnamefont
  {Hern{\'a}ndez-Lobato}}, \bibinfo {author} {\bibfnamefont {B.}~\bibnamefont
  {S{\'a}nchez-Lengeling}}, \bibinfo {author} {\bibfnamefont {D.}~\bibnamefont
  {Sheberla}}, \bibinfo {author} {\bibfnamefont {J.}~\bibnamefont
  {Aguilera-Iparraguirre}}, \bibinfo {author} {\bibfnamefont {T.~D.}\
  \bibnamefont {Hirzel}}, \bibinfo {author} {\bibfnamefont {R.~P.}\
  \bibnamefont {Adams}},\ and\ \bibinfo {author} {\bibfnamefont
  {A.}~\bibnamefont {Aspuru-Guzik}},\ }\bibfield  {title} {\bibinfo {title}
  {Automatic chemical design using a data-driven continuous representation of
  molecules},\ }\href@noop {} {\bibfield  {journal} {\bibinfo  {journal} {ACS
  central science}\ }\textbf {\bibinfo {volume} {4}},\ \bibinfo {pages} {268}
  (\bibinfo {year} {2018})}\BibitemShut {NoStop}%
\bibitem [{\citenamefont {Eckmann}\ \emph {et~al.}(2022)\citenamefont
  {Eckmann}, \citenamefont {Sun}, \citenamefont {Zhao}, \citenamefont {Feng},
  \citenamefont {Gilson},\ and\ \citenamefont {Yu}}]{eckmann2022limo}%
  \BibitemOpen
  \bibfield  {author} {\bibinfo {author} {\bibfnamefont {P.}~\bibnamefont
  {Eckmann}}, \bibinfo {author} {\bibfnamefont {K.}~\bibnamefont {Sun}},
  \bibinfo {author} {\bibfnamefont {B.}~\bibnamefont {Zhao}}, \bibinfo {author}
  {\bibfnamefont {M.}~\bibnamefont {Feng}}, \bibinfo {author} {\bibfnamefont
  {M.~K.}\ \bibnamefont {Gilson}},\ and\ \bibinfo {author} {\bibfnamefont
  {R.}~\bibnamefont {Yu}},\ }\bibfield  {title} {\bibinfo {title} {Limo: Latent
  inceptionism for targeted molecule generation},\ }\href@noop {} {\bibfield
  {journal} {\bibinfo  {journal} {arXiv preprint arXiv:2206.09010}\ } (\bibinfo
  {year} {2022})}\BibitemShut {NoStop}%
\bibitem [{\citenamefont {Bengio}\ \emph {et~al.}(2021)\citenamefont {Bengio},
  \citenamefont {Jain}, \citenamefont {Korablyov}, \citenamefont {Precup},\
  and\ \citenamefont {Bengio}}]{bengio2021flow}%
  \BibitemOpen
  \bibfield  {author} {\bibinfo {author} {\bibfnamefont {E.}~\bibnamefont
  {Bengio}}, \bibinfo {author} {\bibfnamefont {M.}~\bibnamefont {Jain}},
  \bibinfo {author} {\bibfnamefont {M.}~\bibnamefont {Korablyov}}, \bibinfo
  {author} {\bibfnamefont {D.}~\bibnamefont {Precup}},\ and\ \bibinfo {author}
  {\bibfnamefont {Y.}~\bibnamefont {Bengio}},\ }\bibfield  {title} {\bibinfo
  {title} {Flow network based generative models for non-iterative diverse
  candidate generation},\ }\href@noop {} {\bibfield  {journal} {\bibinfo
  {journal} {Advances in Neural Information Processing Systems}\ }\textbf
  {\bibinfo {volume} {34}},\ \bibinfo {pages} {27381} (\bibinfo {year}
  {2021})}\BibitemShut {NoStop}%
\bibitem [{\citenamefont {Roy}\ \emph {et~al.}(2023)\citenamefont {Roy},
  \citenamefont {Bacon}, \citenamefont {Pal},\ and\ \citenamefont
  {Bengio}}]{roy2023goal}%
  \BibitemOpen
  \bibfield  {author} {\bibinfo {author} {\bibfnamefont {J.}~\bibnamefont
  {Roy}}, \bibinfo {author} {\bibfnamefont {P.-L.}\ \bibnamefont {Bacon}},
  \bibinfo {author} {\bibfnamefont {C.}~\bibnamefont {Pal}},\ and\ \bibinfo
  {author} {\bibfnamefont {E.}~\bibnamefont {Bengio}},\ }\bibfield  {title}
  {\bibinfo {title} {Goal-conditioned gflownets for controllable
  multi-objective molecular design},\ }\href@noop {} {\bibfield  {journal}
  {\bibinfo  {journal} {arXiv preprint arXiv:2306.04620}\ } (\bibinfo {year}
  {2023})}\BibitemShut {NoStop}%
\bibitem [{\citenamefont {Vignac}\ \emph {et~al.}(2022)\citenamefont {Vignac},
  \citenamefont {Krawczuk}, \citenamefont {Siraudin}, \citenamefont {Wang},
  \citenamefont {Cevher},\ and\ \citenamefont {Frossard}}]{vignac2022digress}%
  \BibitemOpen
  \bibfield  {author} {\bibinfo {author} {\bibfnamefont {C.}~\bibnamefont
  {Vignac}}, \bibinfo {author} {\bibfnamefont {I.}~\bibnamefont {Krawczuk}},
  \bibinfo {author} {\bibfnamefont {A.}~\bibnamefont {Siraudin}}, \bibinfo
  {author} {\bibfnamefont {B.}~\bibnamefont {Wang}}, \bibinfo {author}
  {\bibfnamefont {V.}~\bibnamefont {Cevher}},\ and\ \bibinfo {author}
  {\bibfnamefont {P.}~\bibnamefont {Frossard}},\ }\bibfield  {title} {\bibinfo
  {title} {Digress: Discrete denoising diffusion for graph generation},\
  }\href@noop {} {\bibfield  {journal} {\bibinfo  {journal} {arXiv preprint
  arXiv:2209.14734}\ } (\bibinfo {year} {2022})}\BibitemShut {NoStop}%
\bibitem [{\citenamefont {Xu}\ \emph {et~al.}(2022)\citenamefont {Xu},
  \citenamefont {Yu}, \citenamefont {Song}, \citenamefont {Shi}, \citenamefont
  {Ermon},\ and\ \citenamefont {Tang}}]{xu2022geodiff}%
  \BibitemOpen
  \bibfield  {author} {\bibinfo {author} {\bibfnamefont {M.}~\bibnamefont
  {Xu}}, \bibinfo {author} {\bibfnamefont {L.}~\bibnamefont {Yu}}, \bibinfo
  {author} {\bibfnamefont {Y.}~\bibnamefont {Song}}, \bibinfo {author}
  {\bibfnamefont {C.}~\bibnamefont {Shi}}, \bibinfo {author} {\bibfnamefont
  {S.}~\bibnamefont {Ermon}},\ and\ \bibinfo {author} {\bibfnamefont
  {J.}~\bibnamefont {Tang}},\ }\bibfield  {title} {\bibinfo {title} {Geodiff: A
  geometric diffusion model for molecular conformation generation},\
  }\href@noop {} {\bibfield  {journal} {\bibinfo  {journal} {arXiv preprint
  arXiv:2203.02923}\ } (\bibinfo {year} {2022})}\BibitemShut {NoStop}%
\bibitem [{\citenamefont {Guimaraes}\ \emph {et~al.}(2017)\citenamefont
  {Guimaraes}, \citenamefont {Sanchez-Lengeling}, \citenamefont {Outeiral},
  \citenamefont {Farias},\ and\ \citenamefont
  {Aspuru-Guzik}}]{guimaraes2017objective}%
  \BibitemOpen
  \bibfield  {author} {\bibinfo {author} {\bibfnamefont {G.~L.}\ \bibnamefont
  {Guimaraes}}, \bibinfo {author} {\bibfnamefont {B.}~\bibnamefont
  {Sanchez-Lengeling}}, \bibinfo {author} {\bibfnamefont {C.}~\bibnamefont
  {Outeiral}}, \bibinfo {author} {\bibfnamefont {P.~L.~C.}\ \bibnamefont
  {Farias}},\ and\ \bibinfo {author} {\bibfnamefont {A.}~\bibnamefont
  {Aspuru-Guzik}},\ }\bibfield  {title} {\bibinfo {title} {Objective-reinforced
  generative adversarial networks (organ) for sequence generation models},\
  }\href@noop {} {\bibfield  {journal} {\bibinfo  {journal} {arXiv preprint
  arXiv:1705.10843}\ } (\bibinfo {year} {2017})}\BibitemShut {NoStop}%
\bibitem [{\citenamefont {You}\ \emph {et~al.}(2018)\citenamefont {You},
  \citenamefont {Liu}, \citenamefont {Ying}, \citenamefont {Pande},\ and\
  \citenamefont {Leskovec}}]{you2018graph}%
  \BibitemOpen
  \bibfield  {author} {\bibinfo {author} {\bibfnamefont {J.}~\bibnamefont
  {You}}, \bibinfo {author} {\bibfnamefont {B.}~\bibnamefont {Liu}}, \bibinfo
  {author} {\bibfnamefont {Z.}~\bibnamefont {Ying}}, \bibinfo {author}
  {\bibfnamefont {V.}~\bibnamefont {Pande}},\ and\ \bibinfo {author}
  {\bibfnamefont {J.}~\bibnamefont {Leskovec}},\ }\bibfield  {title} {\bibinfo
  {title} {Graph convolutional policy network for goal-directed molecular graph
  generation},\ }\href@noop {} {\bibfield  {journal} {\bibinfo  {journal}
  {Advances in neural information processing systems}\ }\textbf {\bibinfo
  {volume} {31}} (\bibinfo {year} {2018})}\BibitemShut {NoStop}%
\bibitem [{\citenamefont {Kong}\ \emph {et~al.}(2023)\citenamefont {Kong},
  \citenamefont {Pang}, \citenamefont {Han},\ and\ \citenamefont
  {Wu}}]{kong2023molecule}%
  \BibitemOpen
  \bibfield  {author} {\bibinfo {author} {\bibfnamefont {D.}~\bibnamefont
  {Kong}}, \bibinfo {author} {\bibfnamefont {B.}~\bibnamefont {Pang}}, \bibinfo
  {author} {\bibfnamefont {T.}~\bibnamefont {Han}},\ and\ \bibinfo {author}
  {\bibfnamefont {Y.}~\bibnamefont {Wu}},\ }\bibfield  {title} {\bibinfo
  {title} {Molecule design by latent space energy-based modeling and gradual
  distribution shifting},\ }\bibfield  {journal} {\bibinfo  {journal}
  {Conference on Uncertainty in Artificial Intelligence}\ }\href
  {https://doi.org/10.48550/arXiv.2306.14902} {10.48550/arXiv.2306.14902}
  (\bibinfo {year} {2023})\BibitemShut {NoStop}%
\bibitem [{\citenamefont {Nigam}\ \emph
  {et~al.}(2022{\natexlab{a}})\citenamefont {Nigam}, \citenamefont {Pollice},\
  and\ \citenamefont {Aspuru-Guzik}}]{nigam2022parallel}%
  \BibitemOpen
  \bibfield  {author} {\bibinfo {author} {\bibfnamefont {A.}~\bibnamefont
  {Nigam}}, \bibinfo {author} {\bibfnamefont {R.}~\bibnamefont {Pollice}},\
  and\ \bibinfo {author} {\bibfnamefont {A.}~\bibnamefont {Aspuru-Guzik}},\
  }\bibfield  {title} {\bibinfo {title} {Parallel tempered genetic algorithm
  guided by deep neural networks for inverse molecular design},\ }\href@noop {}
  {\bibfield  {journal} {\bibinfo  {journal} {Digital Discovery}\ }\textbf
  {\bibinfo {volume} {1}},\ \bibinfo {pages} {390} (\bibinfo {year}
  {2022}{\natexlab{a}})}\BibitemShut {NoStop}%
\bibitem [{\citenamefont {Trabucco}\ \emph {et~al.}(2022)\citenamefont
  {Trabucco}, \citenamefont {Geng}, \citenamefont {Kumar},\ and\ \citenamefont
  {Levine}}]{trabucco2022design}%
  \BibitemOpen
  \bibfield  {author} {\bibinfo {author} {\bibfnamefont {B.}~\bibnamefont
  {Trabucco}}, \bibinfo {author} {\bibfnamefont {X.}~\bibnamefont {Geng}},
  \bibinfo {author} {\bibfnamefont {A.}~\bibnamefont {Kumar}},\ and\ \bibinfo
  {author} {\bibfnamefont {S.}~\bibnamefont {Levine}},\ }\bibfield  {title}
  {\bibinfo {title} {Design-bench: Benchmarks for data-driven offline
  model-based optimization},\ }\href@noop {} {\bibfield  {journal} {\bibinfo
  {journal} {arXiv preprint arXiv:2202.08450}\ } (\bibinfo {year}
  {2022})}\BibitemShut {NoStop}%
\bibitem [{\citenamefont {Linder}\ and\ \citenamefont
  {Seelig}(2021)}]{linder2021fast}%
  \BibitemOpen
  \bibfield  {author} {\bibinfo {author} {\bibfnamefont {J.}~\bibnamefont
  {Linder}}\ and\ \bibinfo {author} {\bibfnamefont {G.}~\bibnamefont
  {Seelig}},\ }\bibfield  {title} {\bibinfo {title} {Fast activation
  maximization for molecular sequence design},\ }\href@noop {} {\bibfield
  {journal} {\bibinfo  {journal} {BMC bioinformatics}\ }\textbf {\bibinfo
  {volume} {22}},\ \bibinfo {pages} {1} (\bibinfo {year} {2021})}\BibitemShut
  {NoStop}%
\bibitem [{\citenamefont {Shen}\ \emph {et~al.}(2021)\citenamefont {Shen},
  \citenamefont {Krenn}, \citenamefont {Eppel},\ and\ \citenamefont
  {Aspuru-Guzik}}]{shen2021deep}%
  \BibitemOpen
  \bibfield  {author} {\bibinfo {author} {\bibfnamefont {C.}~\bibnamefont
  {Shen}}, \bibinfo {author} {\bibfnamefont {M.}~\bibnamefont {Krenn}},
  \bibinfo {author} {\bibfnamefont {S.}~\bibnamefont {Eppel}},\ and\ \bibinfo
  {author} {\bibfnamefont {A.}~\bibnamefont {Aspuru-Guzik}},\ }\bibfield
  {title} {\bibinfo {title} {Deep molecular dreaming: Inverse machine learning
  for de-novo molecular design and interpretability with surjective
  representations},\ }\href@noop {} {\bibfield  {journal} {\bibinfo  {journal}
  {Machine Learning: Science and Technology}\ }\textbf {\bibinfo {volume}
  {2}},\ \bibinfo {pages} {03LT02} (\bibinfo {year} {2021})}\BibitemShut
  {NoStop}%
\bibitem [{\citenamefont {Ramakrishnan}\ \emph {et~al.}(2014)\citenamefont
  {Ramakrishnan}, \citenamefont {Dral}, \citenamefont {Rupp},\ and\
  \citenamefont {Von~Lilienfeld}}]{ramakrishnan2014quantum}%
  \BibitemOpen
  \bibfield  {author} {\bibinfo {author} {\bibfnamefont {R.}~\bibnamefont
  {Ramakrishnan}}, \bibinfo {author} {\bibfnamefont {P.~O.}\ \bibnamefont
  {Dral}}, \bibinfo {author} {\bibfnamefont {M.}~\bibnamefont {Rupp}},\ and\
  \bibinfo {author} {\bibfnamefont {O.~A.}\ \bibnamefont {Von~Lilienfeld}},\
  }\bibfield  {title} {\bibinfo {title} {Quantum chemistry structures and
  properties of 134 kilo molecules},\ }\href@noop {} {\bibfield  {journal}
  {\bibinfo  {journal} {Scientific data}\ }\textbf {\bibinfo {volume} {1}},\
  \bibinfo {pages} {1} (\bibinfo {year} {2014})}\BibitemShut {NoStop}%
\bibitem [{\citenamefont {Wu}\ \emph {et~al.}(2018)\citenamefont {Wu},
  \citenamefont {Ramsundar}, \citenamefont {Feinberg}, \citenamefont {Gomes},
  \citenamefont {Geniesse}, \citenamefont {Pappu}, \citenamefont {Leswing},\
  and\ \citenamefont {Pande}}]{wu2018moleculenet}%
  \BibitemOpen
  \bibfield  {author} {\bibinfo {author} {\bibfnamefont {Z.}~\bibnamefont
  {Wu}}, \bibinfo {author} {\bibfnamefont {B.}~\bibnamefont {Ramsundar}},
  \bibinfo {author} {\bibfnamefont {E.~N.}\ \bibnamefont {Feinberg}}, \bibinfo
  {author} {\bibfnamefont {J.}~\bibnamefont {Gomes}}, \bibinfo {author}
  {\bibfnamefont {C.}~\bibnamefont {Geniesse}}, \bibinfo {author}
  {\bibfnamefont {A.~S.}\ \bibnamefont {Pappu}}, \bibinfo {author}
  {\bibfnamefont {K.}~\bibnamefont {Leswing}},\ and\ \bibinfo {author}
  {\bibfnamefont {V.}~\bibnamefont {Pande}},\ }\bibfield  {title} {\bibinfo
  {title} {Moleculenet: a benchmark for molecular machine learning},\
  }\href@noop {} {\bibfield  {journal} {\bibinfo  {journal} {Chemical science}\
  }\textbf {\bibinfo {volume} {9}},\ \bibinfo {pages} {513} (\bibinfo {year}
  {2018})}\BibitemShut {NoStop}%
\bibitem [{\citenamefont {Nakata}\ and\ \citenamefont
  {Shimazaki}(2017)}]{nakata2017pubchemqc}%
  \BibitemOpen
  \bibfield  {author} {\bibinfo {author} {\bibfnamefont {M.}~\bibnamefont
  {Nakata}}\ and\ \bibinfo {author} {\bibfnamefont {T.}~\bibnamefont
  {Shimazaki}},\ }\bibfield  {title} {\bibinfo {title} {Pubchemqc project: a
  large-scale first-principles electronic structure database for data-driven
  chemistry},\ }\href@noop {} {\bibfield  {journal} {\bibinfo  {journal}
  {Journal of chemical information and modeling}\ }\textbf {\bibinfo {volume}
  {57}},\ \bibinfo {pages} {1300} (\bibinfo {year} {2017})}\BibitemShut
  {NoStop}%
\bibitem [{\citenamefont {Hachmann}\ \emph {et~al.}(2011)\citenamefont
  {Hachmann}, \citenamefont {Olivares-Amaya}, \citenamefont {Atahan-Evrenk},
  \citenamefont {Amador-Bedolla}, \citenamefont {S{\'a}nchez-Carrera},
  \citenamefont {Gold-Parker}, \citenamefont {Vogt}, \citenamefont {Brockway},\
  and\ \citenamefont {Aspuru-Guzik}}]{hachmann2011harvard}%
  \BibitemOpen
  \bibfield  {author} {\bibinfo {author} {\bibfnamefont {J.}~\bibnamefont
  {Hachmann}}, \bibinfo {author} {\bibfnamefont {R.}~\bibnamefont
  {Olivares-Amaya}}, \bibinfo {author} {\bibfnamefont {S.}~\bibnamefont
  {Atahan-Evrenk}}, \bibinfo {author} {\bibfnamefont {C.}~\bibnamefont
  {Amador-Bedolla}}, \bibinfo {author} {\bibfnamefont {R.~S.}\ \bibnamefont
  {S{\'a}nchez-Carrera}}, \bibinfo {author} {\bibfnamefont {A.}~\bibnamefont
  {Gold-Parker}}, \bibinfo {author} {\bibfnamefont {L.}~\bibnamefont {Vogt}},
  \bibinfo {author} {\bibfnamefont {A.~M.}\ \bibnamefont {Brockway}},\ and\
  \bibinfo {author} {\bibfnamefont {A.}~\bibnamefont {Aspuru-Guzik}},\
  }\bibfield  {title} {\bibinfo {title} {The harvard clean energy project:
  large-scale computational screening and design of organic photovoltaics on
  the world community grid},\ }\href@noop {} {\bibfield  {journal} {\bibinfo
  {journal} {The Journal of Physical Chemistry Letters}\ }\textbf {\bibinfo
  {volume} {2}},\ \bibinfo {pages} {2241} (\bibinfo {year} {2011})}\BibitemShut
  {NoStop}%
\bibitem [{\citenamefont {Nigam}\ \emph
  {et~al.}(2022{\natexlab{b}})\citenamefont {Nigam}, \citenamefont {Pollice},
  \citenamefont {Tom}, \citenamefont {Jorner}, \citenamefont {Thiede},
  \citenamefont {Kundaje},\ and\ \citenamefont
  {Aspuru-Guzik}}]{nigam2022tartarus}%
  \BibitemOpen
  \bibfield  {author} {\bibinfo {author} {\bibfnamefont {A.}~\bibnamefont
  {Nigam}}, \bibinfo {author} {\bibfnamefont {R.}~\bibnamefont {Pollice}},
  \bibinfo {author} {\bibfnamefont {G.}~\bibnamefont {Tom}}, \bibinfo {author}
  {\bibfnamefont {K.}~\bibnamefont {Jorner}}, \bibinfo {author} {\bibfnamefont
  {L.~A.}\ \bibnamefont {Thiede}}, \bibinfo {author} {\bibfnamefont
  {A.}~\bibnamefont {Kundaje}},\ and\ \bibinfo {author} {\bibfnamefont
  {A.}~\bibnamefont {Aspuru-Guzik}},\ }\bibfield  {title} {\bibinfo {title}
  {Tartarus: A benchmarking platform for realistic and practical inverse
  molecular design},\ }\href@noop {} {\bibfield  {journal} {\bibinfo  {journal}
  {arXiv preprint arXiv:2209.12487}\ } (\bibinfo {year}
  {2022}{\natexlab{b}})}\BibitemShut {NoStop}%
\bibitem [{\citenamefont {Tripp}\ and\ \citenamefont
  {Hern{\'a}ndez-Lobato}(2023)}]{tripp2023genetic}%
  \BibitemOpen
  \bibfield  {author} {\bibinfo {author} {\bibfnamefont {A.}~\bibnamefont
  {Tripp}}\ and\ \bibinfo {author} {\bibfnamefont {J.~M.}\ \bibnamefont
  {Hern{\'a}ndez-Lobato}},\ }\bibfield  {title} {\bibinfo {title} {Genetic
  algorithms are strong baselines for molecule generation},\ }\href@noop {}
  {\bibfield  {journal} {\bibinfo  {journal} {arXiv preprint arXiv:2310.09267}\
  } (\bibinfo {year} {2023})}\BibitemShut {NoStop}%
\bibitem [{\citenamefont {Wildman}\ and\ \citenamefont
  {Crippen}(1999)}]{wildman1999prediction}%
  \BibitemOpen
  \bibfield  {author} {\bibinfo {author} {\bibfnamefont {S.~A.}\ \bibnamefont
  {Wildman}}\ and\ \bibinfo {author} {\bibfnamefont {G.~M.}\ \bibnamefont
  {Crippen}},\ }\bibfield  {title} {\bibinfo {title} {Prediction of
  physicochemical parameters by atomic contributions},\ }\href@noop {}
  {\bibfield  {journal} {\bibinfo  {journal} {Journal of chemical information
  and computer sciences}\ }\textbf {\bibinfo {volume} {39}},\ \bibinfo {pages}
  {868} (\bibinfo {year} {1999})}\BibitemShut {NoStop}%
\bibitem [{\citenamefont {Lipinski}(2001)}]{lipinski2001lombardo}%
  \BibitemOpen
  \bibfield  {author} {\bibinfo {author} {\bibfnamefont {C.}~\bibnamefont
  {Lipinski}},\ }\bibfield  {title} {\bibinfo {title} {a, lombardo, f., dominy,
  bw \& feeney, pj experimental and computational approaches to estimate
  solubility and permeability in drug discovery and development settings},\
  }\href@noop {} {\bibfield  {journal} {\bibinfo  {journal} {Adv. Drug Deliv.
  Rev}\ }\textbf {\bibinfo {volume} {46}},\ \bibinfo {pages} {00129} (\bibinfo
  {year} {2001})}\BibitemShut {NoStop}%
\bibitem [{\citenamefont {Landrum}\ \emph {et~al.}(2024)\citenamefont
  {Landrum}, \citenamefont {Tosco}, \citenamefont {Kelley}, \citenamefont
  {Rodriguez}, \citenamefont {Cosgrove}, \citenamefont {sriniker},
  \citenamefont {Vianello}, \citenamefont {gedeck}, \citenamefont
  {NadineSchneider}, \citenamefont {Jones}, \citenamefont {Kawashima},
  \citenamefont {Nealschneider}, \citenamefont {Dalke}, \citenamefont {Cole},
  \citenamefont {Swain}, \citenamefont {Turk}, \citenamefont {Savelev},
  \citenamefont {Vaucher}, \citenamefont {Wójcikowski}, \citenamefont {Take},
  \citenamefont {Scalfani}, \citenamefont {Walker}, \citenamefont {Ujihara},
  \citenamefont {Probst}, \citenamefont {guillaume godin}, \citenamefont
  {Pahl}, \citenamefont {Lehtivarjo}, \citenamefont {Berenger}, \citenamefont
  {jasondbiggs},\ and\ \citenamefont {strets123}}]{greg_landrum_2024_10893044}%
  \BibitemOpen
  \bibfield  {author} {\bibinfo {author} {\bibfnamefont {G.}~\bibnamefont
  {Landrum}}, \bibinfo {author} {\bibfnamefont {P.}~\bibnamefont {Tosco}},
  \bibinfo {author} {\bibfnamefont {B.}~\bibnamefont {Kelley}}, \bibinfo
  {author} {\bibfnamefont {R.}~\bibnamefont {Rodriguez}}, \bibinfo {author}
  {\bibfnamefont {D.}~\bibnamefont {Cosgrove}}, \bibinfo {author} {\bibnamefont
  {sriniker}}, \bibinfo {author} {\bibfnamefont {R.}~\bibnamefont {Vianello}},
  \bibinfo {author} {\bibnamefont {gedeck}}, \bibinfo {author} {\bibnamefont
  {NadineSchneider}}, \bibinfo {author} {\bibfnamefont {G.}~\bibnamefont
  {Jones}}, \bibinfo {author} {\bibfnamefont {E.}~\bibnamefont {Kawashima}},
  \bibinfo {author} {\bibfnamefont {D.}~\bibnamefont {Nealschneider}}, \bibinfo
  {author} {\bibfnamefont {A.}~\bibnamefont {Dalke}}, \bibinfo {author}
  {\bibfnamefont {B.}~\bibnamefont {Cole}}, \bibinfo {author} {\bibfnamefont
  {M.}~\bibnamefont {Swain}}, \bibinfo {author} {\bibfnamefont
  {S.}~\bibnamefont {Turk}}, \bibinfo {author} {\bibfnamefont {A.}~\bibnamefont
  {Savelev}}, \bibinfo {author} {\bibfnamefont {A.}~\bibnamefont {Vaucher}},
  \bibinfo {author} {\bibfnamefont {M.}~\bibnamefont {Wójcikowski}}, \bibinfo
  {author} {\bibfnamefont {I.}~\bibnamefont {Take}}, \bibinfo {author}
  {\bibfnamefont {V.~F.}\ \bibnamefont {Scalfani}}, \bibinfo {author}
  {\bibfnamefont {R.}~\bibnamefont {Walker}}, \bibinfo {author} {\bibfnamefont
  {K.}~\bibnamefont {Ujihara}}, \bibinfo {author} {\bibfnamefont
  {D.}~\bibnamefont {Probst}}, \bibinfo {author} {\bibnamefont {guillaume
  godin}}, \bibinfo {author} {\bibfnamefont {A.}~\bibnamefont {Pahl}}, \bibinfo
  {author} {\bibfnamefont {J.}~\bibnamefont {Lehtivarjo}}, \bibinfo {author}
  {\bibfnamefont {F.}~\bibnamefont {Berenger}}, \bibinfo {author} {\bibnamefont
  {jasondbiggs}},\ and\ \bibinfo {author} {\bibnamefont {strets123}},\ }\href
  {https://doi.org/10.5281/zenodo.10893044} {\bibinfo {title} {rdkit/rdkit:
  2024\_03\_1 (q1 2024) release}} (\bibinfo {year} {2024})\BibitemShut
  {NoStop}%
\bibitem [{\citenamefont {Irwin}\ \emph {et~al.}(2012)\citenamefont {Irwin},
  \citenamefont {Sterling}, \citenamefont {Mysinger}, \citenamefont {Bolstad},\
  and\ \citenamefont {Coleman}}]{irwin2012zinc}%
  \BibitemOpen
  \bibfield  {author} {\bibinfo {author} {\bibfnamefont {J.~J.}\ \bibnamefont
  {Irwin}}, \bibinfo {author} {\bibfnamefont {T.}~\bibnamefont {Sterling}},
  \bibinfo {author} {\bibfnamefont {M.~M.}\ \bibnamefont {Mysinger}}, \bibinfo
  {author} {\bibfnamefont {E.~S.}\ \bibnamefont {Bolstad}},\ and\ \bibinfo
  {author} {\bibfnamefont {R.~G.}\ \bibnamefont {Coleman}},\ }\bibfield
  {title} {\bibinfo {title} {Zinc: a free tool to discover chemistry for
  biology},\ }\href@noop {} {\bibfield  {journal} {\bibinfo  {journal} {Journal
  of chemical information and modeling}\ }\textbf {\bibinfo {volume} {52}},\
  \bibinfo {pages} {1757} (\bibinfo {year} {2012})}\BibitemShut {NoStop}%
\bibitem [{\citenamefont {Plante}\ and\ \citenamefont
  {Werner}()}]{Plante2018jplogp}%
  \BibitemOpen
  \bibfield  {author} {\bibinfo {author} {\bibfnamefont {J.}~\bibnamefont
  {Plante}}\ and\ \bibinfo {author} {\bibfnamefont {S.}~\bibnamefont
  {Werner}},\ }\bibfield  {title} {\bibinfo {title} {Jplogp: an improved logp
  predictor trained using predicted data},\ }\bibfield  {journal} {\bibinfo
  {journal} {Journal Of Cheminformatics}\ }\textbf {\bibinfo {volume} {10}},\
  \href {https://doi.org/10.1186/s13321-018-0316-5}
  {10.1186/s13321-018-0316-5}\BibitemShut {NoStop}%
\bibitem [{\citenamefont {Xie}\ and\ \citenamefont
  {Grossman}(2018)}]{xie2018crystal}%
  \BibitemOpen
  \bibfield  {author} {\bibinfo {author} {\bibfnamefont {T.}~\bibnamefont
  {Xie}}\ and\ \bibinfo {author} {\bibfnamefont {J.~C.}\ \bibnamefont
  {Grossman}},\ }\bibfield  {title} {\bibinfo {title} {Crystal graph
  convolutional neural networks for an accurate and interpretable prediction of
  material properties},\ }\href@noop {} {\bibfield  {journal} {\bibinfo
  {journal} {Physical review letters}\ }\textbf {\bibinfo {volume} {120}},\
  \bibinfo {pages} {145301} (\bibinfo {year} {2018})}\BibitemShut {NoStop}%
\bibitem [{\citenamefont {Neese}\ \emph {et~al.}(2020)\citenamefont {Neese},
  \citenamefont {Wennmohs}, \citenamefont {Becker},\ and\ \citenamefont
  {Riplinger}}]{neese2020orca}%
  \BibitemOpen
  \bibfield  {author} {\bibinfo {author} {\bibfnamefont {F.}~\bibnamefont
  {Neese}}, \bibinfo {author} {\bibfnamefont {F.}~\bibnamefont {Wennmohs}},
  \bibinfo {author} {\bibfnamefont {U.}~\bibnamefont {Becker}},\ and\ \bibinfo
  {author} {\bibfnamefont {C.}~\bibnamefont {Riplinger}},\ }\bibfield  {title}
  {\bibinfo {title} {The orca quantum chemistry program package},\ }\href@noop
  {} {\bibfield  {journal} {\bibinfo  {journal} {The Journal of chemical
  physics}\ }\textbf {\bibinfo {volume} {152}} (\bibinfo {year}
  {2020})}\BibitemShut {NoStop}%
\bibitem [{\citenamefont {Kingma}\ and\ \citenamefont
  {Ba}(2014)}]{kingma2014adam}%
  \BibitemOpen
  \bibfield  {author} {\bibinfo {author} {\bibfnamefont {D.~P.}\ \bibnamefont
  {Kingma}}\ and\ \bibinfo {author} {\bibfnamefont {J.}~\bibnamefont {Ba}},\
  }\bibfield  {title} {\bibinfo {title} {Adam: A method for stochastic
  optimization},\ }\href {https://arxiv.org/abs/1412.6980v9} {\bibfield
  {journal} {\bibinfo  {journal} {International Conference on Learning
  Representations}\ } (\bibinfo {year} {2014})}\BibitemShut {NoStop}%
\bibitem [{\citenamefont {Resheff}\ \emph {et~al.}(2017)\citenamefont
  {Resheff}, \citenamefont {Mandelbom},\ and\ \citenamefont
  {Weinshall}}]{resheff2017controlling}%
  \BibitemOpen
  \bibfield  {author} {\bibinfo {author} {\bibfnamefont {Y.~S.}\ \bibnamefont
  {Resheff}}, \bibinfo {author} {\bibfnamefont {A.}~\bibnamefont {Mandelbom}},\
  and\ \bibinfo {author} {\bibfnamefont {D.}~\bibnamefont {Weinshall}},\
  }\bibfield  {title} {\bibinfo {title} {Controlling imbalanced error in deep
  learning with the log bilinear loss},\ }in\ \href@noop {} {\emph {\bibinfo
  {booktitle} {First International Workshop on Learning with Imbalanced
  Domains: Theory and Applications}}}\ (\bibinfo {organization} {PMLR},\
  \bibinfo {year} {2017})\ pp.\ \bibinfo {pages} {141--151}\BibitemShut
  {NoStop}%
\end{thebibliography}%

\onecolumngrid
\newpage

\appendix

\begin{center}
\Large Supplemental Information
\end{center}

\section{Limiting the number of bonds}

In this section we present two strategies to prevent generated adjacency matrices to contain atoms that form an unrealistic number of bonds. For the scope of this paper, we consider that no atom should form more than 4 bonds (e.g., a triple bond and a single bond would count as 4 bonds). In other words, the maximum permitted valence for an atom is 4. 

The first strategy adds an additional penalty to the loss function as follows.  
\begin{equation} \label{eq:constraint}
\mathbb{L}_s = \sum_\Omega \mathbf{A}_{ik} \text{ with } \Omega = \{(i,k)|\sum_{k'} a_{ik'} > 4.5\}
\end{equation}
The second strategy prevents gradients in the direction of higher bonding when there are already 4 bonds. After each epoch, $\mathbf{w_{adj}}^{(t)}$ is updated by the optimizer and the adjacency matrix $\mathbf{A}^{(t+1)}$ is computed from $\mathbf{w_{adj}}^{(t+1)}$. Consider the matrix $\mathbf{I}$ mapping the indices of $\mathbf{w_{adj}}$ to the elements of $\mathbf{A}$ such that $\mathbf{A}_{i,j} = [\mathbf{w_{adj}}_{I_{i,j}}^2]_{\text{sloped}}$. We can build the following list of indices:
\begin{equation}
    \mathbf{l} = \{I_{i,j} | \mathbf{A}^{(t+1)}_{i,j} > \mathbf{A}^{(t)}_{i,j} \land  \sum_{k'} \mathbf{A}^{(t+1)}_{i,k'} > 4.5\}
\end{equation}
In essence, $\mathbf{l}$ is a list of $\mathbf{w_{adj}}$ indices for which the corresponding bond (or element) in the adjacency matrix increased in the last epoch \textbf{and} is part of an atom (or row) that has already formed more than 4 bonds. Finally, we set $\mathbf{w_{adj}}^{(t+1)}_{i\in\mathbf{l}} = \mathbf{w_{adj}}^{(t)}_{i}$ manually. The advantage of this method as opposed to modifying the gradient directly is that it can be applied to any optimizer.

Not all indices of $\mathbf{l}$ are necessary to prevent valence higher than 4. For example, if an atom went from forming 1 bond to 5 bonds in one epoch, the increase from 1 bond to 4 bonds is perfectly allowable. In practice, indices are added to $\mathbf{l}$ until the number of bonds of a specific atom (or row) is 4 or less, but not further.  

In principle, either approach is sufficient on its own. However, the first approach makes the optimization more difficult by adding an additional objective and the second approach cannot handle cases where the starting weights create high valences, which is almost always the case when starting from random weights. When starting from existing molecules, only the second approach is used. 

\section{Feature vector construction}

We construct the feature vector with the following function,

\begin{equation}\label{eq:fea}
\mathbb{F}_{ij} = f_j(\sum_k \mathbf{A}_{i,k}) \mathbf{w_{fea}}^{(ij)}
\end{equation}

where $\mathbb{A}_{i,j}$ and $\mathbb{F}_{i,j}$ are the elements of the adjacency matrix and the feature vector respectively. Each column of the one-hot-encoded feature matrix is associated with a specific element, function $f_j$ defined by a bell-shaped function centered around the number of bonds that an element can form (here index $j$ determines that element). It is described by the following expression:
\begin{equation} \label{eq:feature_vec}
f_j(x) = \left\{
\begin{aligned}
c_0 e^{-{\frac{x - b_j}{\sigma^2}}} + c_1, & \text{    if } b_j - x_0 < x < b_j + x_0 \\
\pm 2 \frac{c_0}{\sigma^2} e^{-\frac{x_0}{\sigma^2}} \left(x_0 x \pm x_0^2 - b_j + \sigma^2\right) + c_1, \span \\
& \text{if } \mp x > x_0 \mp b_j \\
\end{aligned} \right.
\end{equation}
Where $x_0 =  \sigma \sqrt{-\text{ln}(c_2)}$, $c_{0,1,2}$ and $\sigma$ are adjustable parameters, $b_j$ is the number of bonds associated with each atom (e.g., H:1, C:4, O:2) and $\mathbf{w_{fea}}$ are the trainable weights of the feature vector.
\begin{figure}
\includegraphics[width=0.5\linewidth]{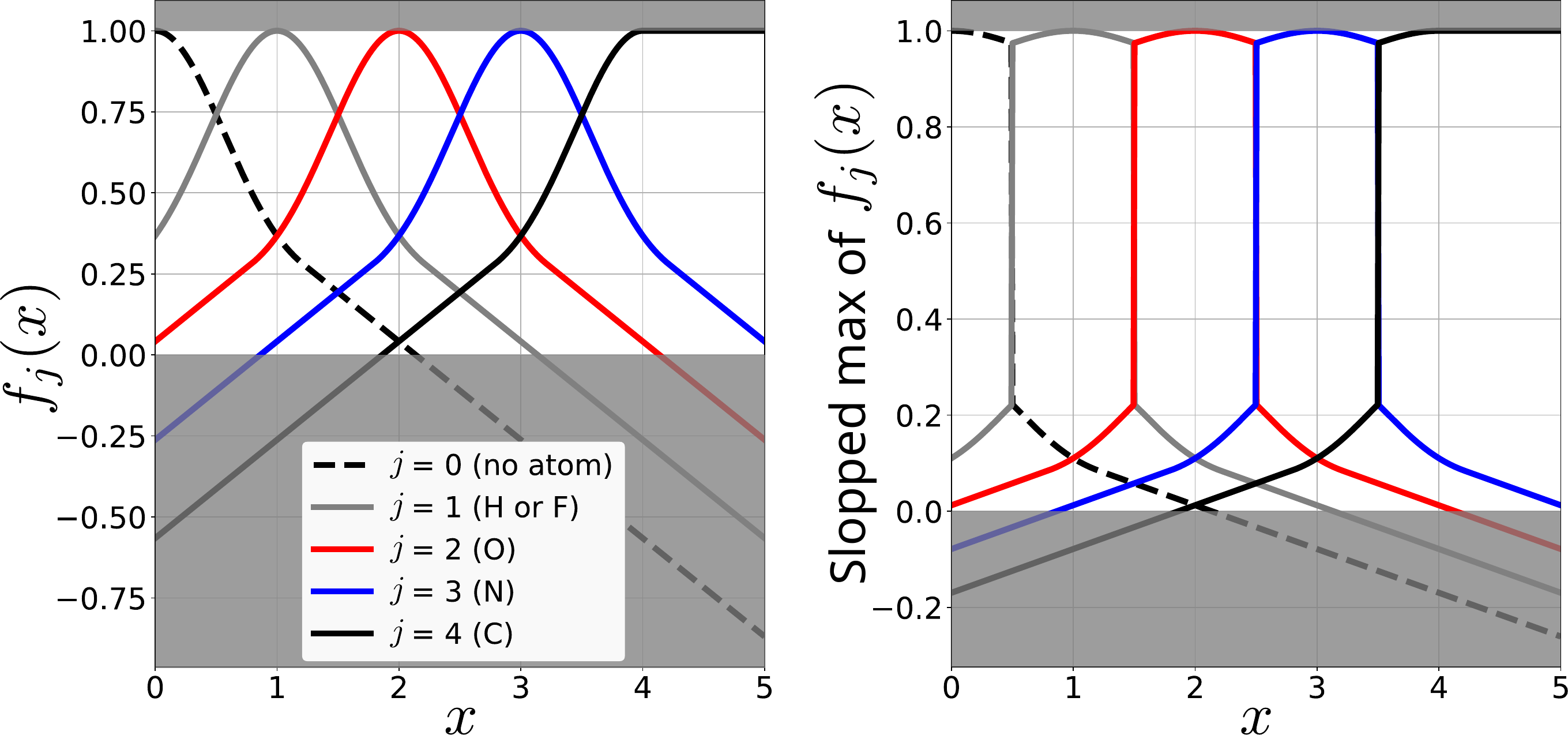}
\caption{\label{fig:feature_vec}
a) Graphical representation of equation \ref{eq:feature_vec} b) Graphical representation of \ref{eq:feature_vec} after applying the sloped maximum function. It shows how the one hot encoding of the feature vector is constructed from the number of bonds ($x$) in the adjacency matrix.}
\end{figure}
 Therefore, if an atom forms 3 bonds for example, only the columns corresponding to elements with a valence of 3 will have sizable values. The shape of $f_j$ for each valance $j$ is plotted in Figure~\ref{fig:feature_vec}. The trainable weights $\mathbf{w_{fea}}$ are only used to differentiate between elements forming the same number of bonds. For example, H and F both form a single bond (valence of 1), in that case $\mathbf{w_{fea}}$ is a single weight that determines if the element is H (0) or F (1). Finally, to ensure that the one-hot encoding are as close as possible to binary with normalized rows, we use a sloped maximum function defined as:

\begin{equation}
    (\text{sloped max } \mathbf{x})_i = \left\{
\begin{aligned}
1 - a(1 - x_i), & \text{ if } x_i = \text{max } \mathbf{x} \\
b x_i, & \text{ otherwise}
\end{aligned} \right. 
\end{equation}

\section{Single component graphs}

With our current implementation, there is no implicit guarantee that the generated graph will be a single molecule. To circumvent this problem, when the stopping criterion is met for the property being optimized, if there is more than one component in the generated graph (i.e., more than one molecule), only the largest component is kept and all other components are removed. The property is then recomputed. If the criterion is still met the generation stops, otherwise, a small amount of random noise is added to the weights to avoid returning to the previous state and the generation continues. 

\section{Details about the energy gap predictor}

The energy gap predictor is a convolutional graph neural network composed of an embedding layer of width 256, 6 convolutional layers, a learned pooling layer and 3 fully connected layers of width 256, 256 and 128. 

The learned pooling layer is described by:
\begin{equation}
    H_\text{pooled} = (\sigma(HW))^T H
\end{equation}
If $B$ is the batch size, $N$ is the maximum number of atoms and $F$ is the number of atomic features, then $H$ is a $B \times N \times F$ hidden layer, $W$ is a $F \times 1$ weight matrix and $H_\text{pooled}$ is a $B \times F$ pooled hidden layer. This is akin to a single transformer block. The activation function $\sigma$ is a sigmoid function for the convolution layer and a softplus function for all other layers.

\begin{figure}
\includegraphics[width=0.5\linewidth] {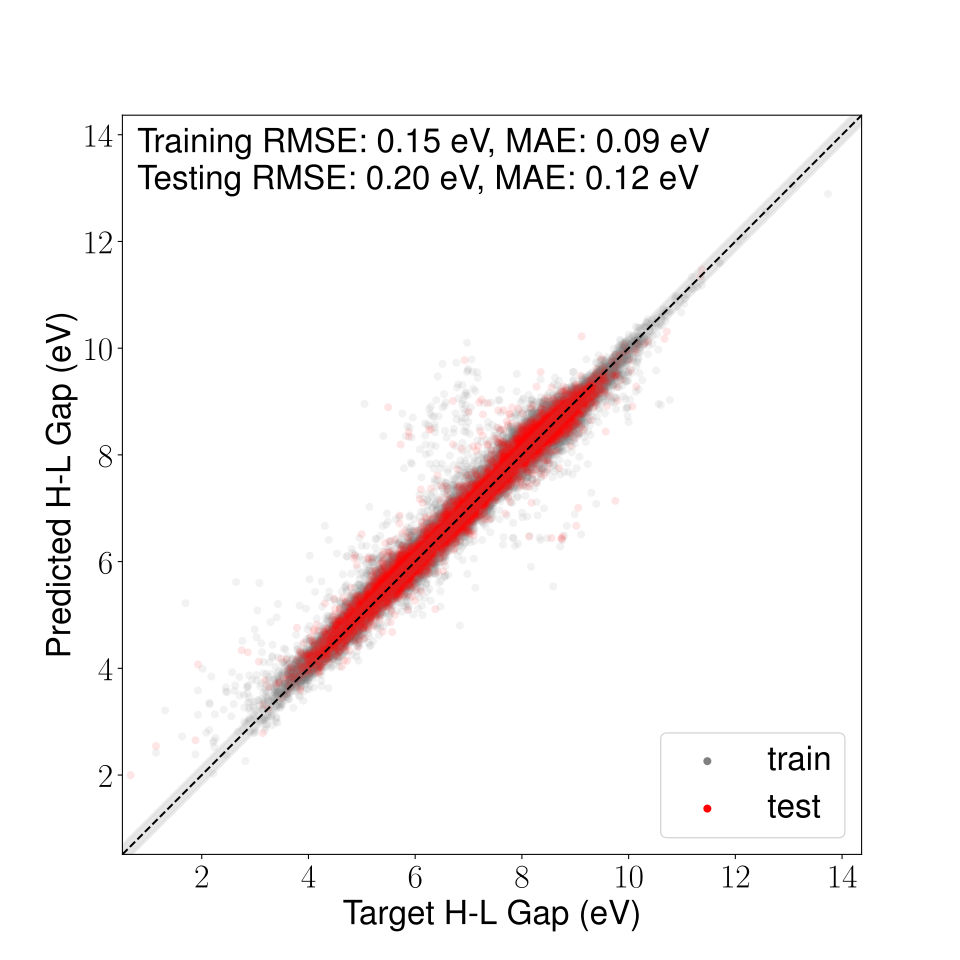} 
\caption{\label{fig:gap_perf}
HOMO-LUMO gap predictor performance}
\end{figure}

During training, convolution layers were batch normalized and a dropout of 10\% was applied before every fully connected layer (including in the convolutions). We chose a maximum number of atoms of 25 and a batch size of 1000. Atoms were described by a one hot encoding of length 5 plus a sixth number representing their number of valence electrons which could be easily calculated from the one hot encoding. Therefore, each batch was composed of a $1000 \times 25 \times 25$ adjacency matrix and a $1000 \times 25 \times 6$ feature vector. A random noise factor with a normal distribution of standard deviation 0.05 was added to both of these matrices at every batch iteration in order to make the network less sensitive to near integers. The model was trained on an 80-20 split of the QM9 dataset for 1000 epochs with the Adam optimizer \cite{kingma2014adam} using the mean absolute error as a loss function. Its prediction accuracy is illustrated in Figure~\ref{fig:gap_perf}.
It is similar to the performance of other models trained on this dataset using the same amount of information (2D graphs). For the purpose of generation, as discussed in the main text, generalizability is what is most important. Here we tuned the hyper parameters by hand to obtain a model as general as possible with acceptable accuracy, without aiming for record-setting values. Full hyper-parameter configuration files are available on our public repository.

\section{Details about the logP predictor (CrippenNet)}

As described in the manuscript, the idea of CrippenNet is to learn the 69 classes of atoms used by Wildman and Crippen \cite{wildman1999prediction} to predict the ``Crippen LogP'' of the RDKit as closely as possible. The 69 classes are broken down into 28 for carbon, 15 for nitrogen, 13 for oxygen, 5 for hydrogen, 3 for sulfur, 1 for fluorine, 1 for chlorine, 1 for bromine, 1 for iodine and 1 for phosphorus. There are other classes of atoms in the original paper, but they are for elements that are not present in our datasets. The challenge here is mainly to correctly classify carbon, nitrogen and oxygen atoms.

CrippenNet is composed of 5 pure convolutional layers (without any weights) that are concatenated into a single feature matrix. The matrix is then fed into 5 different two-layer (width: 100, 28) fully connected networks that predict the classes for carbon, nitrogen, oxygen, hydrogen and sulfur with output layers of width 28, 25, 13 and 5 respectively. Their output is concatenated and multiplied by the feature vector such that atoms can only be classified as the correct element. Elements that have a single class are trivially classified using the feature vector. During training the output for a single molecule is an $N \times 69$ matrix. At inference time, the matrix is multiplied with the $69 \times 1$ vector of individual logP values taken directly from Ref~\cite{wildman1999prediction} to obtain logP for the whole molecule.

\begin{figure}
\includegraphics[width=0.6\linewidth] {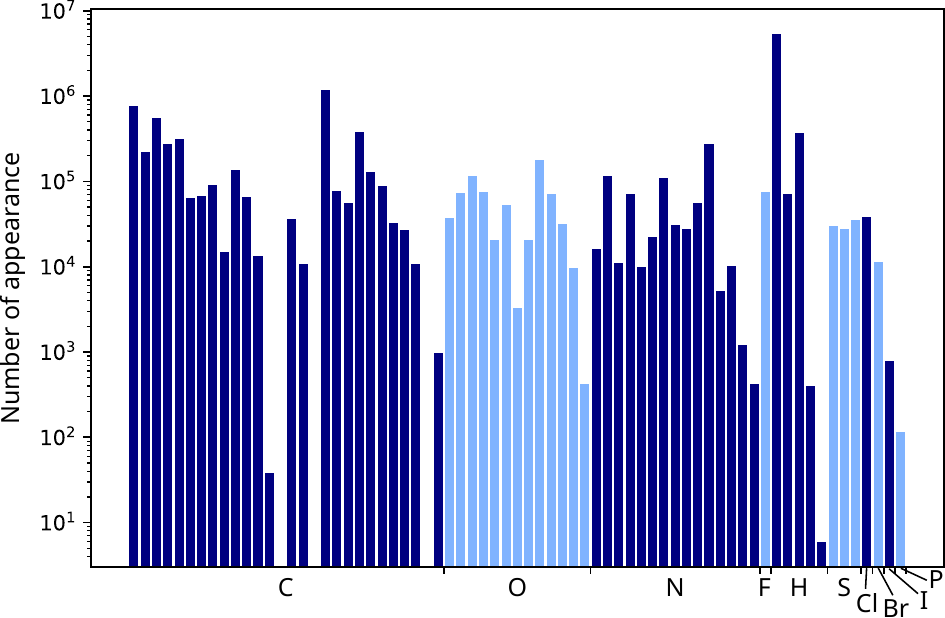}
\caption{\label{fig:class_dist}
Atom class distribution}
\end{figure}

To train this network we computed the actual Wildman \& Crippen classes using the RDKit for all atoms of the 250,000 molecules in the ZINC subset and all atoms of the 133,232 molecules in the QM9 dataset. That creates a total of 11,995,655 examples. The dataset is extremely imbalanced with classes appearing multiple orders of magnitude more than others and some that are never encountered. The first class for hydrogen: hydrocarbons (a hydrogen atom bonded to a carbon) represents 44\% of the dataset. The log distribution of classes is presented in Figure~\ref{fig:class_dist}. 

To deal with this severe imbalance for each batch we randomly sample the classes (with repetition) weighted on the inverse of their occurrence in the dataset. In order to do that we need to first, compute scores for the entire batch and then select the atoms that were randomly picked. This is because each atom's classification depends on the entire molecule that it belongs to. The weights for each epoch are a combination of the inverse of the classes frequency ($f$) and the misclassification rate at the end of the last epoch ($m$). They are given by the following equation:
\begin{equation}
w = \frac{1 + \alpha m}{f+\epsilon},  
\end{equation}

Where $\alpha$ is an adjustable parameter and $epsilon$ are there to avoid infinite values when classes never occur. In the limit of zero classification, the weights are equal to the inverse of the class frequencies. We chose to sample such that there are about 100 examples per class per batch with 100 batches per epoch. Looking at Figure~\ref{fig:class_dist} and with manual tuning of these parameters, we find this is a good compromise between undersampling classes that are very common and oversampling uncommon ones.

\begin{figure}
\includegraphics[width=0.5\linewidth] {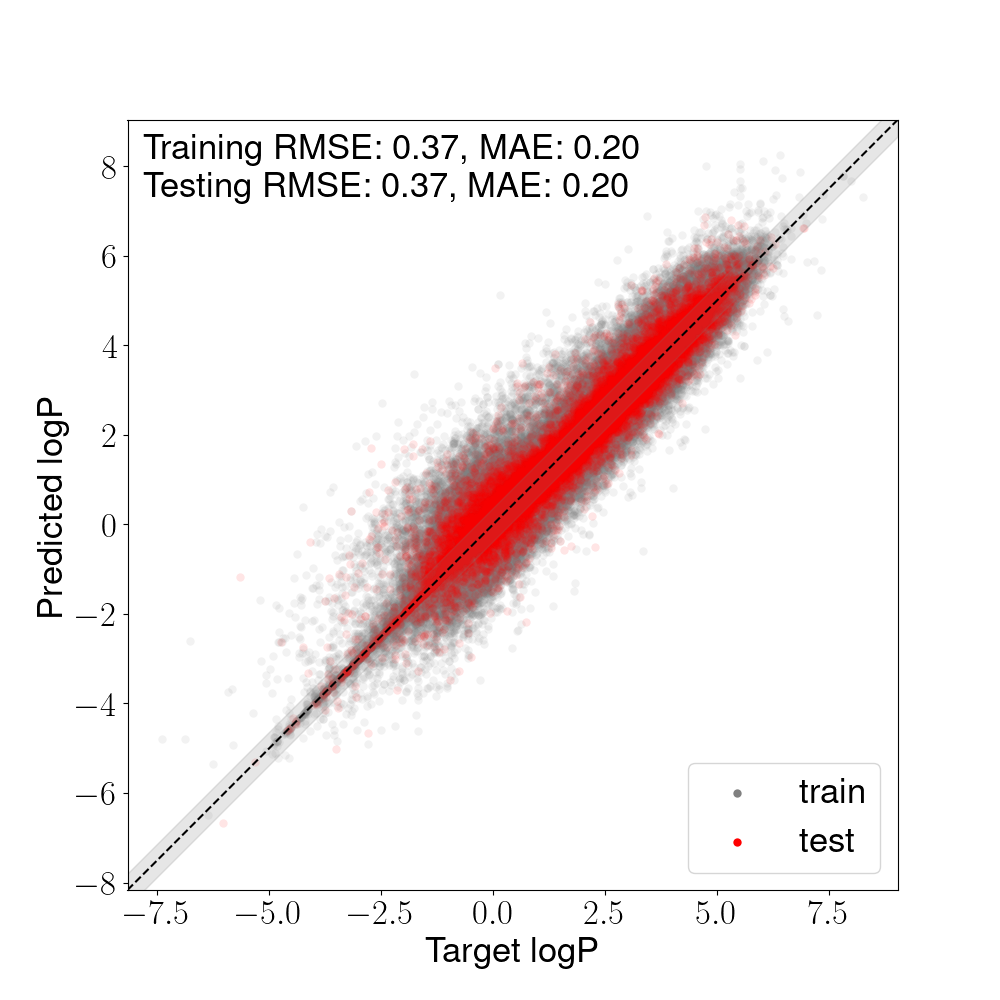} 
\caption{\label{fig:crippenNet}
LogP predictor performance}
\end{figure}

We trained this network for 1000 epochs with a weight decay of $10^{-6}$ using the Adam optimizer and the negative log likelihood loss. We attained a final training accuracy of 98.2\% which translates into a training and testing mean absolute error (MAE) of 0.20~eV. The performance of the model is illustrated in Figure~\ref{fig:crippenNet}.
Note that the train and test MAE are the same, indicating that the model has good generalizability. It is also interesting to note that 38\% of molecules were predicted with an absolute error of less than 0.02.

The remaining challenge here is that some errors are more costly than others. A single misclassification can lead to large differences in logP depending on the classes involved. For example, the minimum atomic logP for carbons is -0.82 whereas the maximum is 0.54, confusing these two classes leads to an error of 1.36 even if only one atom was given the wrong class. Methods like the ones proposed in Ref~\cite{resheff2017controlling} could be used to alleviate this problem.

\section{Details about energy gap targeting}

\begin{figure}
\includegraphics[width=1.0\linewidth]{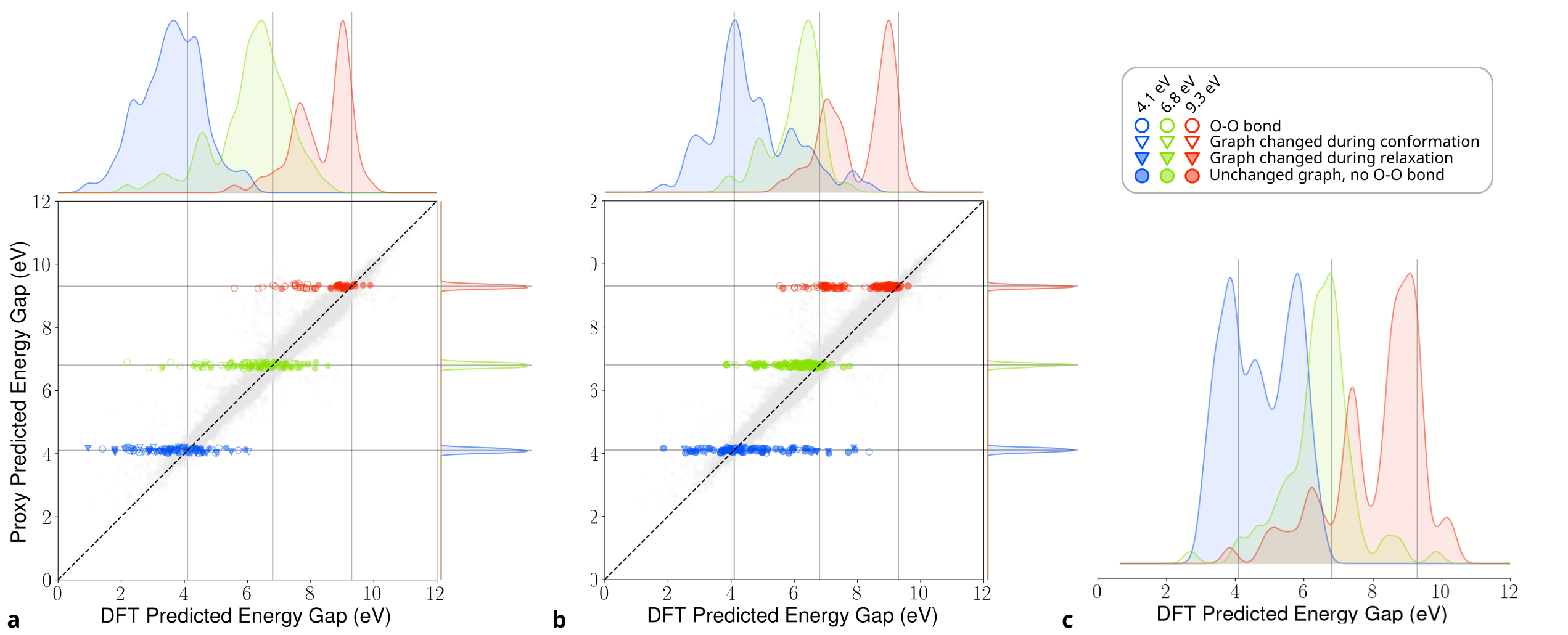}
\caption{\label{fig:gen_others}
Generated molecules HOMO-LUMO gap, DFT and proxy predictions. Generated molecules are overlaid on the proxy model performance on the QM9 dataset (test + train). \textbf{a} DIDGen starting from random noise, \textbf{b} JANUS using our proxy model, \textbf{c} JANUS with DFT}
\end{figure}

The molecular representation was initialized either completely randomly or starting from existing molecules in QM9 and adding a small amount of random noise to their representations. We find that the later gives slightly better results because the proxy model performs better on more QM9-like molecules. Table~\ref{tab:gap_detailed} is a reproduction of Table~\ref{tab:gap} with an additional row for the random generation and Figure~\ref{fig:gen_others}~\textbf{a} show the performance of DIDgen at generating molecules with a specific energy gap starting from random noise.
\begin{table*}
\begin{ruledtabular}
\caption{\label{tab:gap_detailed} Reproduction of Table~\ref{tab:gap} adding DIDgen generation from random noise} 
\begin{tabular}{c|cccc|cccc|cccc}
Method & \multicolumn{4}{c|}{4.1~eV} & \multicolumn{4}{c|}{6.8~eV} & \multicolumn{4}{c}{9.3~eV} \\
 & $n_\text{calcs}$ & $n_{\pm0.5}$ & MAE & Div. & $n_\text{calcs}$ & $n_{\pm0.5}$ & MAE & Div. & $n_\text{calcs}$ & $n_{\pm0.5}$ & MAE & Div.  \\
\colrule
QM9                     &       & 3.4\%         &               & 0.87          &       & 24\%          &               & 0.89          &       & 6.0\% &       & 0.84  \\
\colrule
JANUS (DFT)             & 197   & 24 (12.2\%)            & 0.96          & 0.79          & 392   & 42 (10.7\%)            & 0.92          & 0.80          & 484   & 26 (5.4\%)    & 1.28  & 0.81  \\
JANUS (Proxy)           & 100   & 36            & 1.05          & 0.86          & 100   & 46            & \textbf{0.80} & 0.82          & 100   & \textbf{37}    & 1.24  & 0.81  \\
DIDgen                  & 100   & \textbf{46}            & \textbf{0.81} & \textbf{0.91} & 100   & \textbf{50}   & 0.83          & \textbf{0.90} & 100   & 34    & \textbf{0.83}  & \textbf{0.83}  \\
DIDgen (random start)   & 100   & 38  & 0.87          & 0.89          & 100   & 38            & 0.98          & 0.88          & 100   & 32    & 0.96  & 0.79  \\
\end{tabular}
\end{ruledtabular}
\end{table*}

We used JANUS to maximize the following target: $1/(E_\text{gap} - t + 0.001)$ where $E_\text{gap}$ was evaluated with DFT. We let the algorithm run for 24h on a 40-CPU node. We used a population size of 100 with a maximum of 10 generations. We purposefully limited the compute time such that the comparison with the other methods would be more relevant in terms of computation (see $n_\text{calcs}$ table~\ref{tab:gap}). For the small band gap target, this was not enough to complete more than 2 generations. The distribution of energy gaps for the molecules generated using this method is presented in Figure~\ref{fig:gen_others}~\textbf{c}. 

We also ran JANUS as a basic genetic algorithm (without the neural network) using the proxy to evaluate $E_\text{gap}$. In this case, since evaluations were fast, we used a population size of 1000 and ran it for 10 generations and used the 100 molecules closest to the target in the final population for DFT evaluation. The distribution of energy gaps and the predicted vs. DFT-measured energy gaps are presented in Figure~\ref{fig:gen_others}~\textbf{b}.

In the case of large energy gaps, only 65 (62, random start) of the 100 generated molecules were unique and are represented in Figure~\ref{fig:gen}, Figure~\ref{fig:gen_others}, Table~\ref{tab:gap} and Table~\ref{tab:gap_detailed}, i.e., they are not double counted. Small HOMO-LUMO gap molecules tend to be more complex--intuitively, the more orbitals there are, the higher the chances that the HOMO and the LUMO lay close to each other--which could explain why more molecules are unstable for that objective. Conversely, large gap molecules tend to be simpler, which makes it harder to generate diverse molecules with large gaps and explains why the distribution for that task is skewed asymmetrically towards smaller gaps.

\end{document}